\documentclass[a4paper, 10pt, conference, compsocconf]{IEEEtran}

\usepackage{graphics} 
\usepackage{epsfig} 
\usepackage{mathptmx} 
\usepackage{times} 
\usepackage{amsmath} 
\usepackage{amssymb}  
\usepackage{flexisym}
\usepackage{subcaption}
\usepackage{multirow}
\usepackage{float}

\usepackage{url}
\usepackage{xcolor}

\DeclareMathAlphabet{\pazocal}{OMS}{zplm}{m}{n}

\newcommand{\etal}{\textit{et al}.}

\begin{document}
	\IEEEoverridecommandlockouts\pubid{\makebox[\columnwidth]{978-1-5386-2335-0/18/\$31.00~\copyright{}2018 IEEE \hfill}
		\hspace{\columnsep}\makebox[\columnwidth]{ }}
%
\title{High-Quality Facial Photo-Sketch Synthesis Using Multi-Adversarial Networks}
\author{
	\IEEEauthorblockN{
		Lidan Wang,
		Vishwanath A. Sindagi, Vishal M. Patel }
	\IEEEauthorblockA{    Rutgers, The State University of New Jersey \\
		94 Brett Road, Piscataway, NJ 08854\\
		Email:  \{lidan.wang, vishwanath.sindagi, vishal.m.patel\}@rutgers.edu}}


%


\maketitle

\begin{abstract}
Synthesizing face sketches from real photos and its inverse have many applications. However, photo/sketch synthesis remains a challenging problem due to the fact that photo and sketch have different characteristics. In this work, we consider this task as an image-to-image translation problem and explore the recently  popular generative models (GANs) to generate high-quality realistic photos from sketches and sketches from photos. Recent GAN-based methods have shown promising results on image-to-image translation problems and photo-to-sketch synthesis in particular, however, they are known to have limited abilities in generating high-resolution realistic images. To this end, we propose a novel synthesis framework called Photo-Sketch Synthesis using Multi-Adversarial Networks,  (PS\textsuperscript{2}-MAN) that iteratively generates low resolution to high resolution images in an adversarial way. The hidden layers of the generator are supervised to first generate lower resolution images followed by implicit refinement in the network to generate higher resolution images. Furthermore, since photo-sketch synthesis is a coupled/paired translation problem, we leverage the pair information using CycleGAN framework. Both Image Quality Assessment (IQA) and Photo-Sketch Matching experiments are conducted to demonstrate the superior performance of our framework in comparison to existing state-of-the-art solutions. Code available at:\\ \url{https://github.com/lidan1/PhotoSketchMAN}.
\end{abstract}


%
\IEEEpeerreviewmaketitle

\section{INTRODUCTION}
Research on biometrics has made significant progress in the past few decades and face remains the most commonly studied biometric primarily due to convenience of data collection. In law enforcement and criminal cases, automatic retrieval of photos of suspects from the police mug shot database can enable the authorities to rapidly narrow down potential suspects \cite{wang2009face}. In practice, photos of suspects are usually hard to acquire and it is known that commercial softwares or experienced artists are sought to generate sketches of a suspect based on the description of eyewitness. Other than the applications in security, face photo-sketch synthesis also has several applications in digital entertainment. Photo sketches have also become increasingly popular among the users of smart phones and social networks where sketches are used as profile photos or avatars. Thus, photo-sketch synthesis and matching are important and practical problems.   

Earlier studies on face photo-sketch matching have focused on directly matching photos to sketches and vice versa \cite{wang2009face}. However, due to differences in style and appearance of photo and sketch, it is not practical to  directly perform matching between the two modalities. A common approach to reduce this domain gap between photo and sketch is to perform face photo-synthesis technique prior to matching. Several algorithms are proposed on this topic in the literature. Existing approaches can be generally classified into four categories based on the types of sketches used\cite{peng2016face}: (i) hand-drawn viewed sketch \cite{wang2009face},\cite{zhou2012markov}, (ii) hand-drawn semi-forensic sketch \cite{ouyang2016forgetmenot}, (iii) hand-drawn forensic sketch \cite{klare2011matching, klare2013heterogeneous}, and (iv) software-generated composite sketch \cite{han2013matching}.


\begin{figure}[t!]
	\centering
	\includegraphics[scale=0.2]{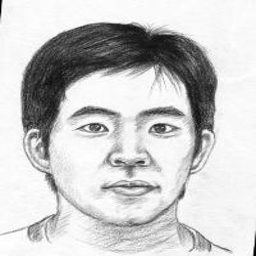}
	\includegraphics[scale=0.532]{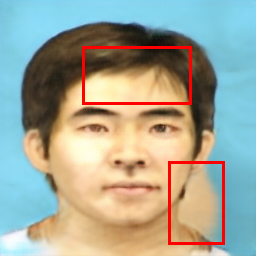}
	\includegraphics[scale=0.2]{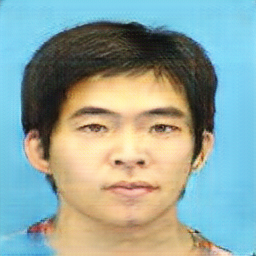}
	\includegraphics[scale=0.2]{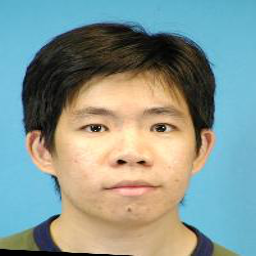}
	\includegraphics[scale=0.2]{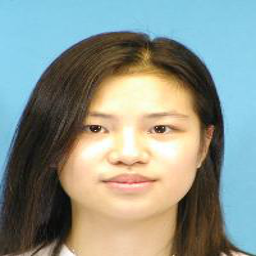}
	\includegraphics[scale=0.2]{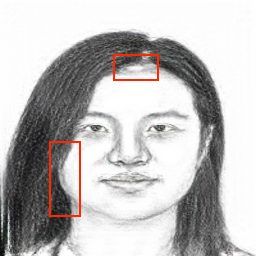}
	\includegraphics[scale=0.2]{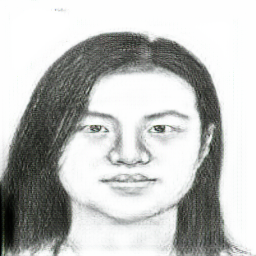}
	\includegraphics[scale=0.2]{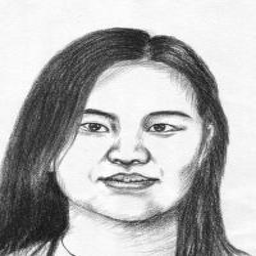} \\
	(a)\hskip50pt(b)\hskip50pt(c)\hskip50pt(d)
	\vskip-5pt
	\caption{Sample results on photo and sketch synthesis. Top Row: Photo Synthesis, Bottom Row: Sketch synthesis. (a) Input Image. (b) Synthesis using single stage adversarial network. (c) Synthesis using multi-stage adversarial network (proposed method). (d) Ground truth. Artifacts in (b) are marked with red rectangles.}
	\label{fig1}
\end{figure} 

Several works have successfully exploited Convolutional Neural Networks (CNNs) to perform different image-to-image translation tasks. Recently, generative models such as Generative Adversarial Networks (GANs)  \cite{goodfellow2014generative} and Variational Auto-Encoders (VAE)  \cite{kingma2013auto,rezende2014stochastic} have been more successful in such tasks due to their powerful generative abilities. In particular, GANs \cite{goodfellow2014generative} have achieved impressive results in image generation, image editing and representation learning \cite{sindagi2017generating,zhang2017image,perera2017in2i,di2017gp,zhang2017joint}. Recent studies also adopt the original method for conditional image generation tasks such as image-to-image translation \cite{isola2016image}. While Isola \etal (\cite{isola2016image}) considered paired data for learning the image-to-image translation, Zhu \etal \cite{zhu2017unpaired} and Yi \etal \cite{yi2017dualgan} separately proposed unsupervised image-to-image translation methods without the use of paired data. Similar to  \cite{isola2016image}, \cite{yi2017dualgan} and \cite{zhu2017unpaired}, in this work, face photo-sketch synthesis is considered as an image-to-image translation task. In fact, Yi \etal \cite{yi2017dualgan} presented some preliminary results specifically for photo-sketch synthesis. On evaluating these methods in detail for our task, it was found that they had limitations in generating higher resolution images (as shown in Fig.\ref{fig1}). As argued in \cite{zhang2016stackgan}, it is difficult to train GANs to generate high-resolution realistic images as they tend to generate images with artifacts. This is attributed to the fact that as pixel space dimension increases, the overlap between natural distribution of images and learned model distribution reduces. To overcome this issue, a novel high-quality face photo-sketch synthesis framework based on GANs is proposed. Since in our task, both photo-to-sketch synthesis and sketch-to-photo synthesis have practical applications, we adopt the recently introduced CycleGAN \cite{zhu2017unpaired} framework. Similar to \cite{zhu2017unpaired}, the proposed method has two generators $G_A$ and $G_B$ which generate sketch from photo and photo from sketch, respectively. In contrast to \cite{zhu2017unpaired}, two major differences can be noted: 1) To address the issue of artifacts in high-resolution image generation, we propose the use of multi-discriminator network. 2) While CycleGAN uses only cycle-consistency loss, we additionally use $L_1$ reconstruction error between generated output and target image. The use of additional loss functions behaves as a regularization during the learning process.

Existing GANs use generators that are constructed similar to encoder-decoder style where the input image is first forwarded through a series of convolutions, non-linearities and max-pooling resulting in lower resolution feature maps which are then forwarded through a series of deconvolutions and non-linearities. Noting that the deconvolutions iteratively learn the weights to upsample the feature maps, this implicit presence of feature maps at different resolutions is leveraged in this work by applying adversarial supervision at every level of resolution. Specifically, the feature maps at every deconvolution layer are convolved using $3\times3$ convolutions to produce outputs at different resolutions (3 in particular). A discriminator network is introduced at every resolution. By doing so, supervision is provided directly to hidden layers of the network which will enable iterative refinement of the feature maps and hence the output image. To summarize, this paper makes the following contributions:
\begin{itemize}
	\item A novel face photo-sketch synthesis framework based on GANs involving multi-adversarial networks where adversarial supervision is provided to hidden layers of the network. 
	\item While \cite{yi2017dualgan} and \cite{sangkloy2016scribbler} present generic adversarial methods to perform image-to-image translation and show some preliminary results on face photo-sketch synthesis, to the best of our knowledge, ours is the first work to study in detail the use of adversarial networks specifically for face photo-sketch synthesis. 	
	\item Detailed experiments are conducted to demonstrate improvements in the synthesis results. Further, ablation studies are conducted to verify the effectiveness of iterative synthesis. 
\end{itemize}


\section{RELATED WORK}
In this section, previous work on face photo-sketch synthesis and generative modeling techniques that are applied for image-to-image translation tasks are reviewed.
\subsection{Face photo-sketch synthesis}
Existing works can be categorized based on multiple factors. Wang \etal \cite{wang2014comprehensive} categorize photo-sketch synthesis methods based on model construction techniques into three main classes: 1) subspace learning-based, 2) sparse representation-based, and 3) Bayesian inference-based approaches. Peng \etal \cite{peng2017superpixel} perform the categorization based on representation strategies and come up with three broad approaches: 1)  holistic image-based, 2) independent local patch-based, and 3) local patch with spatial constraints-based methods. 

Subspace learning based methods involve the use of linear and non-linear subspace methods such as Principal Component Analysis (PCA) and Local Linear Embedding (LLE). Tang and Wang \cite{tang2004face,tang2002face} assume linear mapping between photo and sketch and synthesized the sketch by taking a linear combination of the Eigen vectors of sketch images. Finding that the assumption of linear mapping to be unreasonable, Liu \etal \cite{liu2005nonlinear} proposed a non-linear method based on LLE where they perform a patch-based sketch synthesis. The input photo image is divided into overlapping patches and transformed to corresponding sketch patches using the LLE method. The whole sketch image is then obtained by averaging the overlapping areas between neighboring sketch patches. However, it leads to blurring effect and ignores the neighboring relationships among the patches and thus is unable to take advantage of global structure.  This work was extended by Wang \etal \cite{wang2013heterogeneous}, Gao \etal \cite{gao2012face} and Change \etal \cite{chang2010face} using sparse representation-based techniques. In a different approach, several methods were developed using Bayesian inference techniques. Gao \etal \cite{gao2008face} and Xiao \etal \cite{xiao2009new} employed Hidden Markov Model (HMMs) to model non-linear relationship between sketches and photos. Wang and Tang \cite{wang2009face} proposed Markov Random Field (MRF) based technique to incorporate relationship among neighboring patches. Zhou \etal \cite{zhou2012markov} improved over \cite{wang2009face} by proposing Markov weight fields (MWF) model that is capable of synthesizing new target
patches not existing in the training set. Wang et al. \cite{wang2013transductive} proposed a novel face sketch synthesis method based on transductive learning.

More recently, Peng \etal \cite{peng2016multiple} proposed a multiple representations-based face sketch photo-synthesis method that adaptively combines multiple representations to represent an image patch by combining multiple features from face images processed using multiple filters. Additionally, they employ Markov networks to model the relationship between neighboring patches. Zhang \etal \cite{zhang2016robust} employed a sparse representation-based greedy search strategy to first estimate an initial sketch. Candidate image patches from the initial estimated sketch and the template sketch are then selected using multi-scale features. These candidate patches are refined and assembled to obtain the final sketch which is further enhanced using a cascaded regression strategy. Peng \etal \cite{peng2017superpixel} proposed a superpixel-based synthesis method involving two stage synthesis procedure. Wang \etal \cite{wang2017bayesian} recently proposed the use of Bayesian framework consisting of neighbor selection model and weight computation model. They consider spatial neighboring constraint between adjacent image patches for both models in contrast to existing methods where the adjacency constraint is considered for only one of the models. CNN-based method such as \cite{gao2017composition} and \cite{chen2018face} were proposed recently showing promising results. There is also a recent work on face synthesis from facial attribute \cite{di2017face} applying sketch to photo synthesis as a second stage in their approach. 

\subsection{Image-to-image translation}
In contrast to the traditional methods for photo-sketch synthesis, several researchers have exploited the success of CNNs for synthesis and cross-domain photo-sketch recognition. Face photo-sketch synthesis is considered as an image-to-image translation problem. Zhang \etal \cite{zhang2015end}  proposed an end-to-end fully convolutional network-based photo-sketch synthesis method. 
Several methods have been developed for related tasks such as general sketch synthesis \cite{sangkloy2016scribbler}, photo-caricature translation  \cite{zheng2017photo} and creation of parameterized avatars \cite{wolf2017unsupervised}.

In this work, we explore generative modeling techniques which have been highly successful for several image-to-image translation tasks. GANs  \cite{goodfellow2014generative,zhu2017unpaired} and VAEs \cite{rezende2014stochastic,kingma2013auto} are two recently popular classes of generative techniques. GANs  \cite{goodfellow2014generative} are used to synthesize realistic images by learning the distribution of training images. GANs, motivated by game theory, consist of two competing networks: generator $G$ and discriminator $D$. The goal of GAN is to train $G$ to produce samples from training distribution such that the synthesized samples are indistinguishable from actual distribution by discriminator $D$. In another variant called Conditional GAN , the generator is conditioned on additional variables such as discrete labels, text and images \cite{isola2016image}. Recently, several variants based on original GAN have been proposed for image-to-image translation tasks. Isola \etal \cite{isola2016image} proposed Conditional GANs for several tasks such as labels to street scenes, labels to facades, image colorization, etc. In an another variant, Zhu \etal \cite{zhu2017unpaired} proposed CycleGAN that learns image-to-image translation in an unsupervised fashion. Similar to the above approach, Yi \etal \cite{yi2017dualgan} proposed an unsupervised method to perform translation tasks based on unpaired data.


\begin{figure*}[thpb]
	\centering
	\includegraphics[scale=0.4]{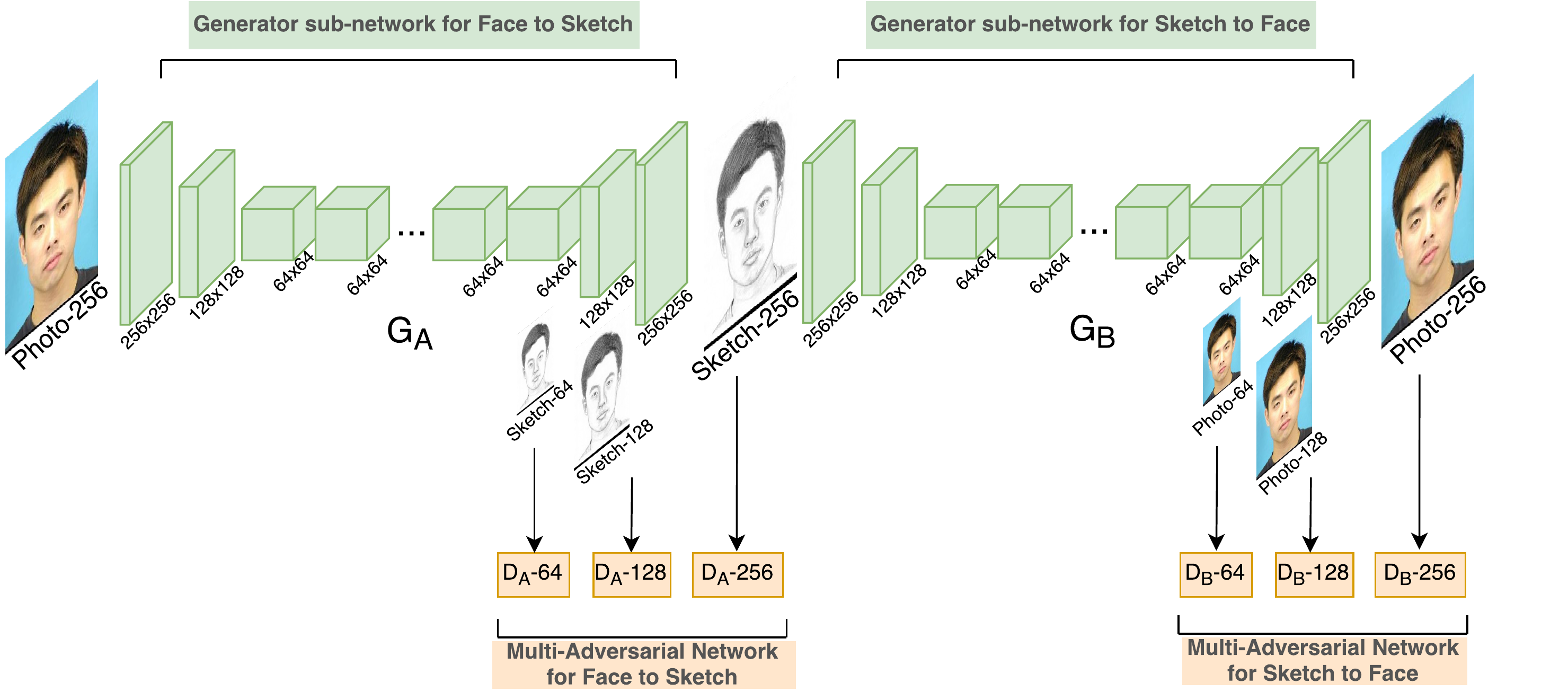}
	\caption{Network structure of the proposed PS\textsuperscript{2}-MAN framework. Adversarial supervision is provided through multiple discriminators at the hidden layers. Note that, in addition to adversarial loss, cycle-consistency and L1-loss are also used to train the network. However, for the purpose of illustration, we show only adversarial loss in this figure. }
	\label{net1}
\end{figure*}

\section{PROPOSED METHOD}
In this section, the problem formulation is presented followed by a detailed description of the proposed method PS\textsuperscript{2}-MAN. Also, details of generator's and discriminator's network architecture are provided. 
\subsection{Formulation}
Given a dataset ($\mathcal{D}$) consisting of a set of face photo-sketch pairs represented by $\{(A_i,B_i)\}_{i=1}^N$, the goal of photo-sketch synthesis is to learn two functions: (1) B\textprime=$f_{ps}(A)$ that represents photo (A) to sketch (B) synthesis and (2) A\textprime=$f_{sp}(B)$ that represents sketch (B) to photo (A) synthesis. In this work, we consider this problem as an image-to-image translation task. Since both forward (photo to sketch) and inverse (sketch to photo) transformations are of equal practical importance, this problem can be easily accommodated into the CycleGAN \cite{zhu2017unpaired} framework. Similar to \cite{zhu2017unpaired}, the proposed method consists of two generator sub-networks $G_A$ and $G_B$ which transform from photo to sketch and from sketch to photo, respectively. $G_A$ takes in a real face photo image $R_A$ as input and produces synthesized (fake) sketch $F_B$ as output. The aim of $G_B$  is to transform sketch to photo, hence,  it should transform $F_B$ back to input photo itself, which we represent as $Rec_A$ here. Thus, the general process can be expressed as:
\begin{equation}\label{eq1}
	F_B = G_A( R_A), \\\
	Rec_A = G_B(F_B).
\end{equation} 
Similarly, sketch to photo generation can be expressed as:
\begin{equation}\label{eq2}
	F_A = G_B( R_B), \\\
	Rec_B= G_A(F_A),
\end{equation} 
where $R_B$, $F_A$ and $Rec_B$ are real sketch, synthesized (fake) photo, and reconstructed sketch from fake photo, respectively. Note that in the following context, the term ``fake" is same as ``synthesized". 

\subsection{Objective}

As in GAN framework \cite{goodfellow2014generative}, the generators ($G_A$ and $G_B$) are trained using adversarial losses that come from discriminator sub-networks. The goal of the generator sub-networks is to produce images that are as realistic as possible so as to fool the discriminator sub-networks, where as the goal of the discriminator sub-networks is to learn to classify between generated and real samples. The use of adversarial loss is known to overcome the issue of blurred outputs that is often encountered when only L1 or L2 loss is minimized \cite{isola2016image}. In theory, GANs can learn a mapping that produce outputs identically distributed as target domain and although generic image-to-image translation GANs have been successful in generating visually appealing results, they tend to produce artifacts in the output (as shown in Fig~\ref{fig1}) which adversely affects the face/sketch matching performance. Hence, it is crucial to generate outputs that are free from artifacts. 

As discussed in \cite{zhang2016stackgan}, these artifacts arise due to  known training instabilities while generating high-resolution images. These instabilities are potentially caused due to the fact that the supports of natural image  distribution and implied model distribution may not overlap in high-dimensional space. The severity of this problem increases with an increase in the  image resolution. Thus, to avoid these artifacts while generating realistic images, we propose a stage-by-stage multi-scale refinement framework by leveraging the implicit presence of features maps of different resolutions in the generator sub-network. Considering that most GAN frameworks have generators similar to encoder-decoder style with a stack of convolutional and max-pooling layers followed by a series of deconvolution layers. The deconvolution layers sequentially upsample the feature maps from lower resolution to higher resolution. Feature maps from every deconvolutional layers are forwarded through $3\times3$ convolutional layer to generate output images at different resolutions. As shown in Fig~\ref{net1}, output images are generated at three resolution levels: $64\times64$, $128\times128$ and $256\times256$ for both generators $G_A$ and $G_B$. Further, three separate discriminator sub-networks are employed to provide adversarial feedback to the generators. By doing so, we are providing supervision directly to hidden layers of the network which will enable implicit iterative refinement of the feature maps resulting in high-quality synthesis. For simplicity, images at different resolutions are represented as: 
$R_{A_i}$, $F_{A_i}$, $Rec_{A_i}$,  $R_{B_i}$, $F_{B_i}$, and $Rec_{B_i}$, where $i = 1,2$ and three corresponds to resolution of $64\times64$, $128\times128 $ and final output size, which is $256 \times 256$.

Thus, as shown in Fig.~\ref{net1}, for a photo image $R_A$,  $G_A$ generates \{$F_{B_1},F_{B_2},F_{B_3}$\} as outputs. Then $F_{B_3}$, which is the output at the last deconvolution layer, is sent as input to $G_{B}$ resulting in three reconstructions \{$Rec_{A_1}$,$Rec_{A_2}$,$Rec_{A_3}$\}. Similarly, for a sketch input, $G_B$ will output \{$F_{A_1}$,$F_{A_2}$,$F_{A_3}$\}.  And $G_A$ will produce \{$Rec_{B_1}$,$Rec_{B_2}$,$Rec_{B_3}$\} by taking $F_{A_3}$ as input. We then add supervision at these different outputs to force outputs to be closer to  targets at different resolution levels. Three discriminators are defined for each generator: $D_{A64}, D_{A128}, D_{A256}$ for $G_A$ and  $D_{B64}, D_{B128}, D_{B256}$ for $G_B$, which are applied on deconvolution layers with resolutions of $64\times64$, $128\times128$ and $256\times256$, respectively. The objective function is expressed as:

\begin{equation}\label{eq4}
	\begin{aligned}
		\pazocal{L}_{GAN_{A_i}} = & \mathbb{E}_{B_i \sim p_{data}(B_i)} [ \log D_{A_i} (B_i)]   \\ 
		& + \mathbb{E}_{A \sim p_{data}(A)} [ \log (1- D_{A_i} (G_A(R_A))_i) ],
	\end{aligned}
\end{equation} 
and
\begin{equation}\label{eq4}
	\begin{aligned}
		\pazocal{L}_{GAN_{B_i}} = & \mathbb{E}_{A_i \sim p_{data}(A_i)} [ \log D_{A_i} (A_i)]   \\ 
		& + \mathbb{E}_{B \sim p_{data}(B)} [ \log (1- D_{B_i} (G_B(R_B))_i) ],
	\end{aligned}
\end{equation} 
where $(G_A(R_A))_i = F_{B_i}$, $(G_B(R_B))_i = F_{A_i}$ and $i = 1,2,3$ corresponds to discriminators at different levels.

To generate images which are as close to target images as possible, we also minimize synthesis error $\pazocal{L}_{syn}$ which is defined as the $L_{1}$ difference between synthesized image and corresponding target image. Similar to adversarial loss,  $\pazocal{L}_{syn}$ is minimized for all three resolution levels and is defined as:
\begin{equation}\label{eq3}
	\begin{aligned}
		&\pazocal{L}_{syn_{A_i}} = \| F_{A_i} -R_{A_i} \|_1 =  \| G_B(R_B)_i -R_{A_i} \|_1 \\
		&\pazocal{L}_{syn_{B_i}} = \| F_{B_i} -R_{B_i} \|_1 = \| G_A(R_A)_i -R_{B_i} \|_1.
	\end{aligned}
\end{equation} 

In spite of using $\pazocal{L}_{syn}$ and the adversarial loss, as discussed in \cite{zhu2017unpaired},  we may have many mappings due to the large capacity of networks. Hence, similar to \cite{zhu2017unpaired}, the network is additionally regularized using forward-backward consistency thereby reducing the space of possible mapping functions. This is achieved by introducing cycle consistency losses at different resolution stages, which are defined as:

\begin{equation}\label{eq4}
	\begin{aligned}
		&\pazocal{L}_{cyc_{A_i}} = \| Rec_{A_i} -R_{A_i} \|_1 =  \| G_B(G_A(R_A))_i -R_{A_i} \|_1   \\ 
		&\pazocal{L}_{cyc_{B_i}} = \| Rec_{B_i} -R_{B_i} \|_1 = \| G_A(G_B(R_B))_i -R_{B_i} \|_1.
	\end{aligned}
\end{equation}

The final objective function is defined as:
\nonumber \begin{equation}\label{eq7}
	\begin{aligned}
		\nonumber \pazocal{L}(G_A, G_B, D_{A},D_{B}) = & \sum_{i=1}^{3}( \pazocal{L}_{GAN_{A_i}}+\pazocal{L}_{GAN_{B_i}}+\lambda_{A_i}\pazocal{L}_{syn_{A_i}}\\
		\nonumber &+\lambda_{B_i} \pazocal{L}_{syn_{B_i}}+\eta_{A_i}\pazocal{L}_{cyc_{A_i}}+\eta_{B_i}\pazocal{L}_{cyc_{B_i}}).  
	\end{aligned}
\end{equation} 

To summarize, the final objective function is constructed using $L_{1}$ error between synthesized and target images, adversarial loss and cycle-consistency loss. $L_{1}$ error enables the network to synthesize images that are closer to the target, however, they often result in blurry images. Adversarial loss overcomes this issue thereby resulting in relatively sharper images. However, the use of adversarial loss at the final stage results in artifacts, which we overcome by providing supervision to the hidden layers. Cycle-consistency loss provides additional regularization while learning the network parameters. 

\subsection{Network Architecture}

The generator sub-networks are constructed using stride-2 convolutions, residual blocks \cite{he2016deep} and fractionally strided convolutional layers. The network configuration is specified as follows:

\noindent {\emph{C7S1-64, C3-128, C3-256, RB256$\times$9, TC64, TC32, C7S1-3,}}
\noindent where, $C7S1-k$ denotes $7 \times 7$ Convolution-BatchNormReLU
layer with $k$ filters and stride 1, $Ck$ denotes a $3 \times 3$ Convolution-BatchNorm-ReLU layer with $k$ filters, and stride 2, $RBk\times m$ denotes $m$ residual block that contains two $3 \times 3$ convolutional layers with the same number of filters on both layers, $TC$ denotes a $3 \times 3$ Transposed-Convolution-BatchNorm-ReLU layer with $k$ filters and stride $\frac{1}{2}$.

The discriminator networks are constructed using $70 \times 70$ PatchGANs \cite{isola2016image} that classify whether $70 \times 70$ overlapping image patches are real or fake. The network configuration is specified as: \emph{C64-C128-C256-C512}, where $Ck$ denotes a $4\times 4$ Convolution-BatchNorm-LeakyReLU layer with
$k$ filters and stride 2. 

\section{EXPERIMENTAL RESULTS}
In this section, experimental settings and evaluation of the proposed method are discussed in detail. We present the qualitative and quantitative results of the proposed method on two popular datasets: CUFS \cite{wang2009face} and CUFSF \cite{wang2009face,zhang2011coupled}. In addition to performance comparison with state-of-the-art methods, ablation studies are conducted to verify the effectiveness of various components of the proposed method.  Both visual and quantitative comparisons are presented in both cases. 

\subsection{Datasets}

The proposed method is evaluated on existing viewed sketches datasets. CUHK Face Sketch database (CUFS) \cite{wang2009face} is a viewed sketch database which includes 188 faces from the Chinese University of Hong Kong (CUHK) student database, 123 faces from the AR database \cite{martinez1998ar}, and 295 faces from the XM2VTS database \cite{messer1999xm2vtsdb}. For each face, there is a sketch drawn by an artist based on a photo taken in a frontal pose under normal lighting condition, and with a neutral expression.  

CUFSF \cite{wang2009face,zhang2011coupled} database includes 1,194 persons from the FERET database \cite{phillips1998feret}. For each person, there is  a face photo with lighting variation and a sketch with shape exaggeration drawn by an artist when viewing this photo \cite{zhang2011coupled}. This dataset is particularly challenging since the photos are taken under different illumination conditions and sketches have shape exaggeration as compared to photos, however, the dataset is closer to forensic sketch scenario.

Both datasets contain facial landmark coordinates which can be easily applied for alignments. There also exist several recent datasets without landmark information, recent face alignment algorithms such as \cite{peng2015piefa} can be applied in the preprocessing stage. 
\subsection{Training Details}
During model training procedure, each input image is resized to the size of $256 \times 256$. Data augmentation is performed on the fly by adding random noise to input images. The network is trained from scratch, similar to the network initialization setup in \cite{zhu2017unpaired}, the learning rate is set to 0.0002 for the first 100 epochs, and linearly decaying down to 0 for next 100 epochs. $\lambda_i$ are all set to 1 and $\eta_i$ are all set to 0.7 in (\ref{eq7}). Weights were initialized from a Gaussian distribution with mean 0 and standard deviation 0.02. The network is trained using the Adam solver \cite{kingma2014adam}. For the CUHK dataset, 188 face-sketch pairs are divided such that 60 pairs are used for training, 28 pairs for validation and 100 pairs for testing. We augmented training set by horizontally flipping images so that training set has 120 images in total. For the CUFSF dataset, 1194 image pairs are divided to 600 for training, 297 for validation and 297 for testing. All images are pre-processed by simply aligning center of the eyes to the fixed position and cropping to the size of $200 \times 250$. 
\begin{figure}[htp!]
	\centering
	\begin{subfigure}[b]{0.08\textwidth}
		\centering
		\includegraphics[width=1\textwidth]{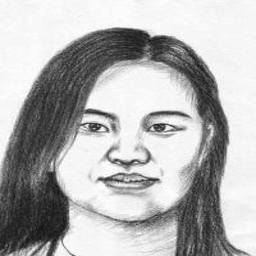}
		\includegraphics[width=1\textwidth]{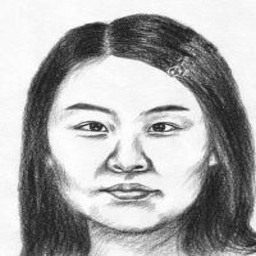}
		\includegraphics[width=1\textwidth]{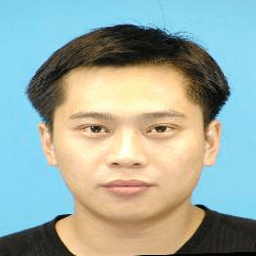}
		\includegraphics[width=1\textwidth]{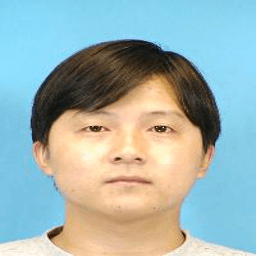}
		\caption{}
	\end{subfigure}
	\begin{subfigure}[b]{0.08\textwidth}
		\centering
		\includegraphics[width=1\textwidth]{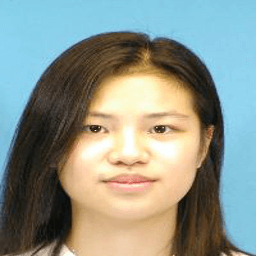}
		\includegraphics[width=1\textwidth]{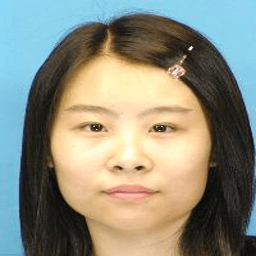}
		\includegraphics[width=1\textwidth]{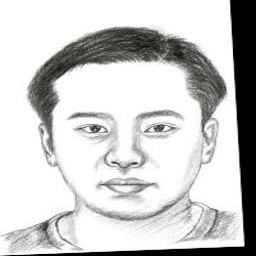}
		\includegraphics[width=1\textwidth]{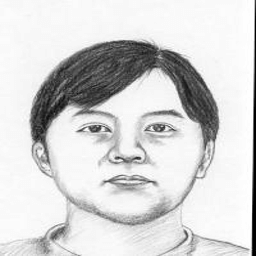}
		\caption{}
	\end{subfigure}
	\begin{subfigure}[b]{0.08\textwidth}
		\centering
		\includegraphics[width=1\textwidth]{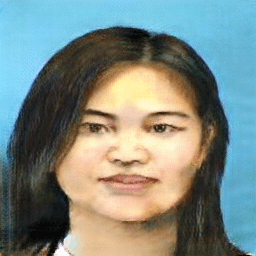}
		\includegraphics[width=1\textwidth]{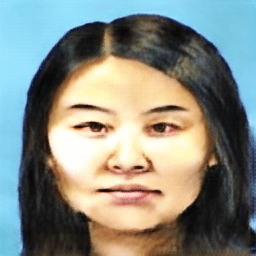}    		\includegraphics[width=1\textwidth]{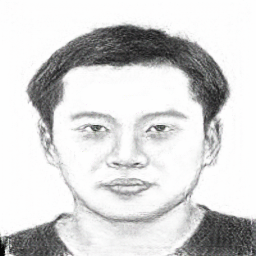}
		\includegraphics[width=1\textwidth]{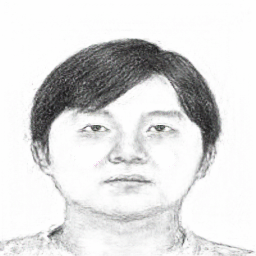}
		\caption{}
	\end{subfigure}
	\begin{subfigure}[b]{0.08\textwidth}
		\centering
		\includegraphics[width=1\textwidth]{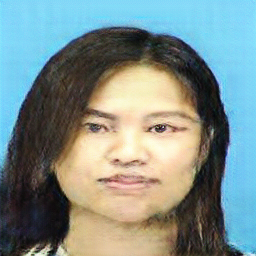}
		\includegraphics[width=1\textwidth]{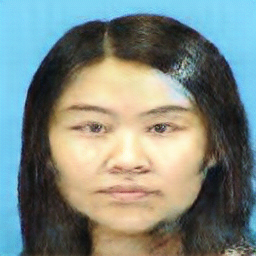}    		\includegraphics[width=1\textwidth]{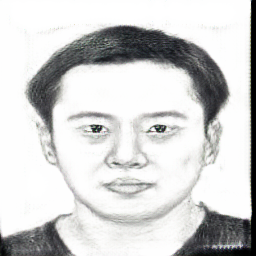}
		\includegraphics[width=1\textwidth]{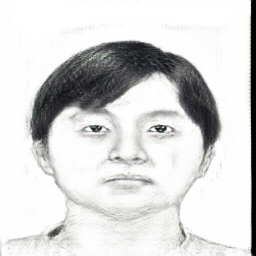}
		\caption{}
	\end{subfigure}
	\begin{subfigure}[b]{0.08\textwidth}
		\centering
		\includegraphics[width=1\textwidth]{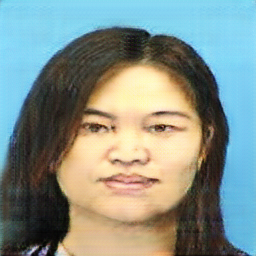}
		\includegraphics[width=1\textwidth]{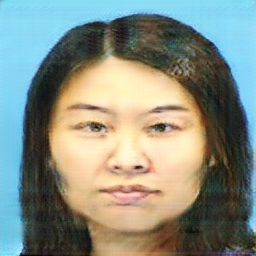}    \includegraphics[width=1\textwidth]{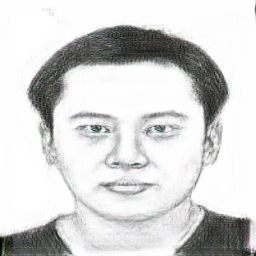}
		\includegraphics[width=1\textwidth]{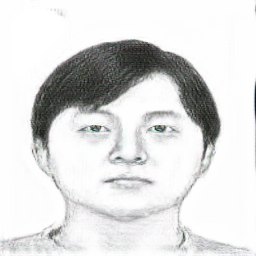}
		\caption{}
	\end{subfigure}
	\vskip-5pt
	\caption{Results of ablation study:  (a) Input. (b) Ground truth. (c) $C-D_{256}$. (d) $C-D_{256,128}$. (e) $C-D_{256,128,64}$. Row 1 and Row 2: Photo synthesis from sketch. Row 3 and Row 4: Sketch synthesis from photo. It can be observed from (e) that the artifacts are minimized and the results are more realistic.} \label{ablation}
\end{figure}

\vskip-15pt

\subsection{Ablation Study}
To demonstrate the advantage of our multi-adversarial network structure over the single adversarial approach, we compare the results of the following network configurations on the CUHK dataset:
\begin{itemize}
	\item$C-D_{256}$: Proposed method with single discriminator at the final resolution level ($256 \times 256$). 
	\item $C-D_{256,128}$: Proposed method with two discriminators at last two resolution levels ($256 \times 256$ and $128 \times 128$). 	
	\item $C-D_{256,128,64}$: Proposed method with two discriminators at three resolution levels ($256 \times 256$,  $128 \times 128$ and $64 \times 64$).
\end{itemize}

Fig. ~\ref{ablation} shows sample results from the above configurations on the CUHK dataset. It can be observed that the performance in terms of visual quality improves as more levels of supervision are added. Similar observations can be made using the quantitative measurements such as SSIM \cite{wang2004image} and FSIM\ \cite{zhang2011fsim}) as shown in Table \ref{Ablation}.

\begin{figure}[t!]
	\centering
	\begin{subfigure}[b]{0.45\textwidth}
		\centering
		\includegraphics[width=0.145\textwidth]{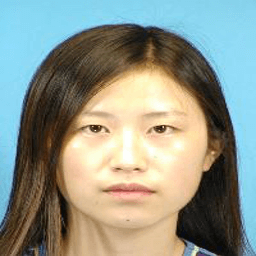}
		\includegraphics[width=0.145\textwidth]{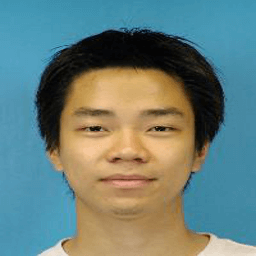}
		\includegraphics[width=0.145\textwidth]{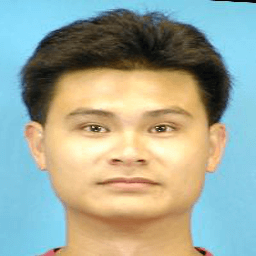}
		\includegraphics[width=0.145\textwidth]{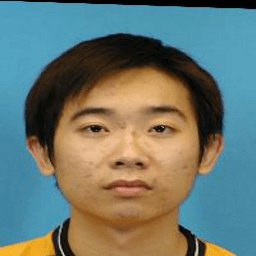}
		\includegraphics[width=0.145\textwidth]{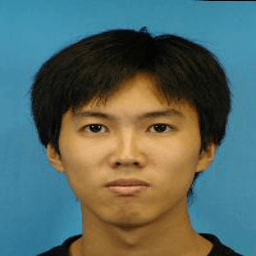}
	\end{subfigure}
	\begin{subfigure}[b]{0.45\textwidth}
		\centering
		\includegraphics[width=0.145\textwidth]{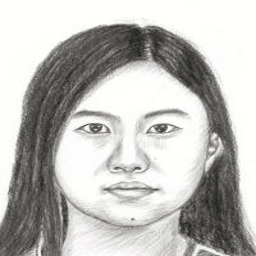}
		\includegraphics[width=0.145\textwidth]{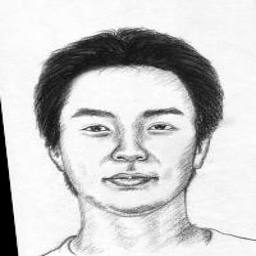}
		\includegraphics[width=0.145\textwidth]{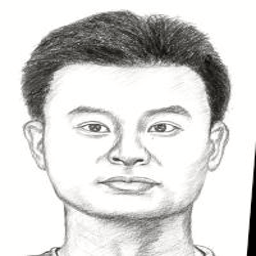}
		\includegraphics[width=0.145\textwidth]{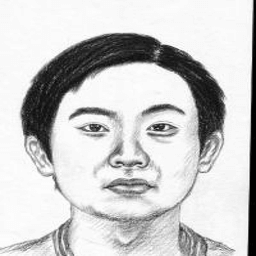}
		\includegraphics[width=0.145\textwidth]{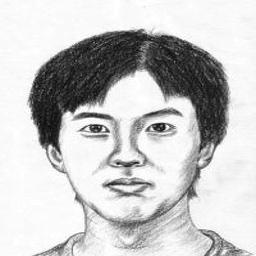}
	\end{subfigure}
	\begin{subfigure}[b]{0.45\textwidth}
		\centering
		\includegraphics[width=0.145\textwidth]{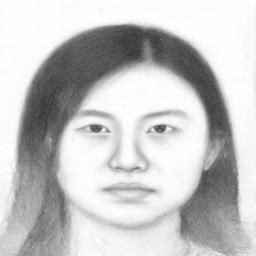}
		\includegraphics[width=0.145\textwidth]{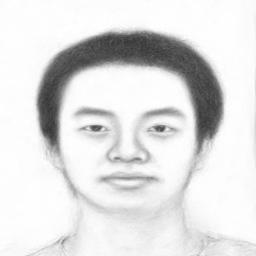}
		\includegraphics[width=0.145\textwidth]{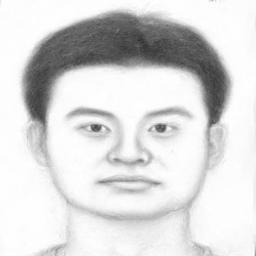}
		\includegraphics[width=0.145\textwidth]{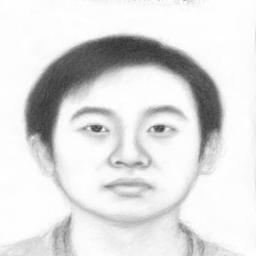}
		\includegraphics[width=0.145\textwidth]{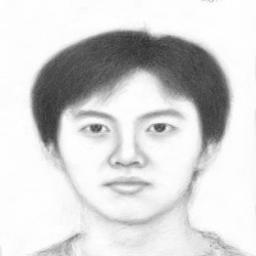}
	\end{subfigure}
	\begin{subfigure}[b]{0.45\textwidth}
		\centering
		\includegraphics[width=0.145\textwidth]{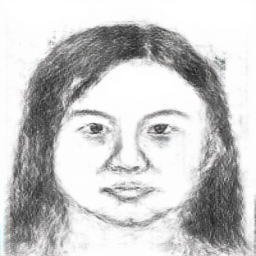}
		\includegraphics[width=0.145\textwidth]{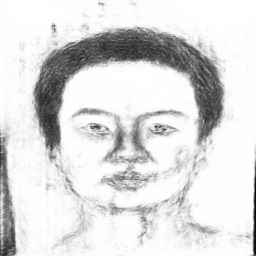}
		\includegraphics[width=0.145\textwidth]{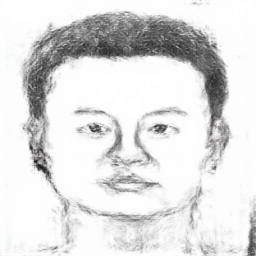}
		\includegraphics[width=0.145\textwidth]{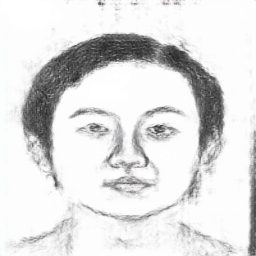}
		\includegraphics[width=0.145\textwidth]{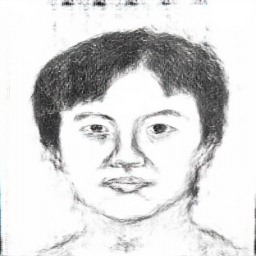}
	\end{subfigure}
	\begin{subfigure}[b]{0.45\textwidth}
		\centering
		\includegraphics[width=0.145\textwidth]{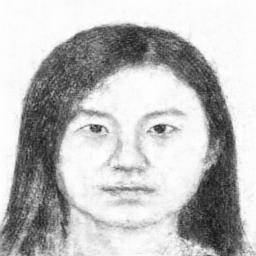}
		\includegraphics[width=0.145\textwidth]{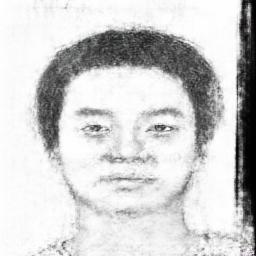}
		\includegraphics[width=0.145\textwidth]{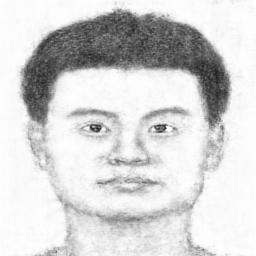}
		\includegraphics[width=0.145\textwidth]{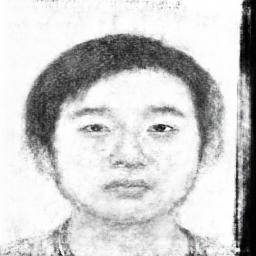}
		\includegraphics[width=0.145\textwidth]{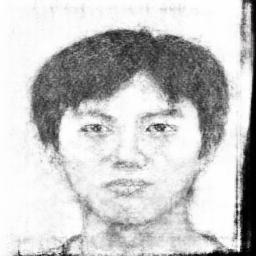}
	\end{subfigure}
	\begin{subfigure}[b]{0.45\textwidth}
		\centering
		\includegraphics[width=0.145\textwidth]{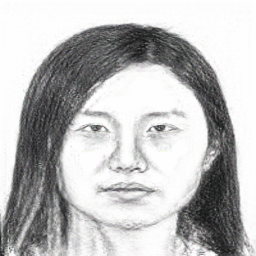}
		\includegraphics[width=0.145\textwidth]{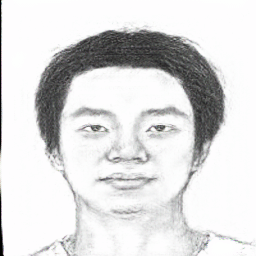}
		\includegraphics[width=0.145\textwidth]{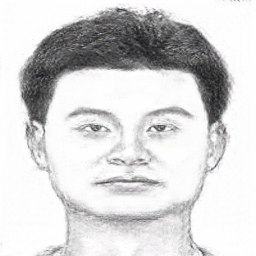}
		\includegraphics[width=0.145\textwidth]{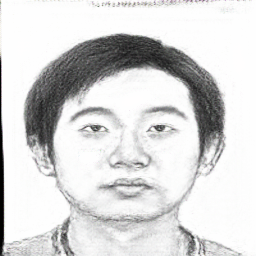}
		\includegraphics[width=0.145\textwidth]{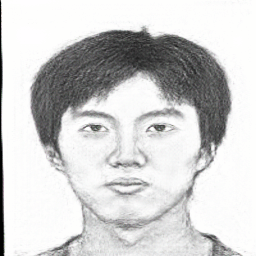}
	\end{subfigure}
	
	\begin{subfigure}[b]{0.45\textwidth}
		\centering
		\includegraphics[width=0.145\textwidth]{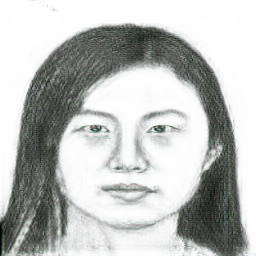}
		\includegraphics[width=0.145\textwidth]{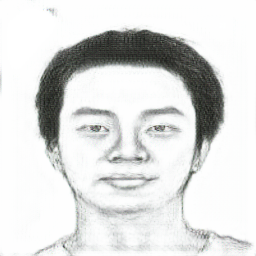}
		\includegraphics[width=0.145\textwidth]{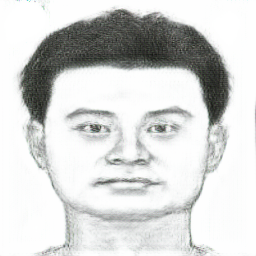}
		\includegraphics[width=0.145\textwidth]{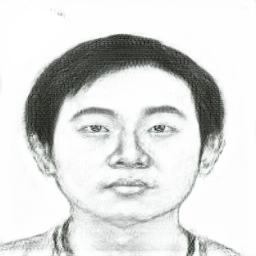}
		\includegraphics[width=0.145\textwidth]{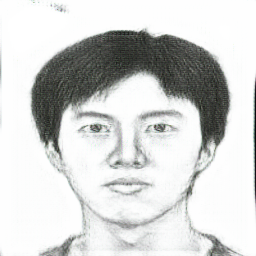}
	\end{subfigure}
	\caption{Comparison of photo to sketch synthesis results on the CUHK dataset. From top to bottom: Input, Ground truth, MrFSPS, Pix2Pix, DualGAN, CycleGAN and PS\textsuperscript{2}-MAN. PS\textsuperscript{2}-MAN has minimal artifacts while generating realistic and sharper images.}\label{cuhkcompare}
\end{figure}

\begin{table}[htp!]
	\caption{\scriptsize{ABLATION STUDY: QUANTITATIVE RESULTS FOR PHOTO AND SKETCH SYNTHESIS FOR DIFFERENT CONFIGURATIONS ON CUHK DATASET}}
	\vskip-12pt	
	\label{Ablation}
	
	\begin{center}
		\resizebox{0.48\textwidth}{!}{%
			\begin{tabular}{|l|c|c|c|c|c|c|}
				\hline
				& $C-D_{256}$ & $C-D_{256,128}$ &  $C-D_{256,128,64}$ \\ \hline
				SSIM (Photo Synthesis)  & 0.7626   & 0.7851          & 0.7915     \\ \hline
				SSIM (Sketch Synthesis) & 0.5991   & 0.6034          & 0.6156     \\ \hline
				FSIM (Photo Synthesis)  & 0.7826   & 0.7920          & 0.8062     \\ \hline
				FSIM (Sketch Synthesis) & 0.7271   & 0.7280          & 0.7361     \\ \hline
				
			\end{tabular}
		}
	\end{center}
	
\end{table}

\begin{figure}[h!]
	\centering
	\begin{subfigure}[b]{0.45\textwidth}
		\centering
		\includegraphics[width=0.145\textwidth]{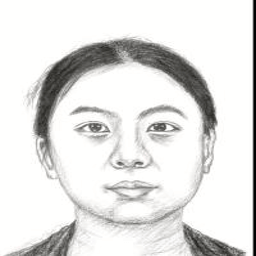}
		\includegraphics[width=0.145\textwidth]{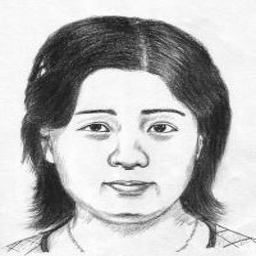}
		\includegraphics[width=0.145\textwidth]{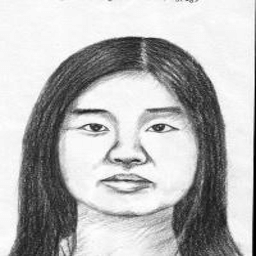}
		\includegraphics[width=0.145\textwidth]{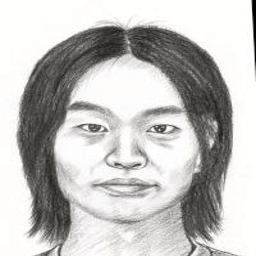}
		\includegraphics[width=0.145\textwidth]{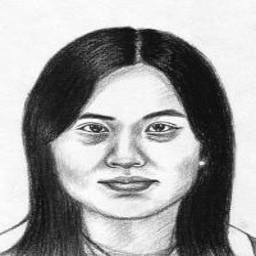}
	\end{subfigure}
	
	\begin{subfigure}[b]{0.45\textwidth}
		\centering
		\includegraphics[width=0.145\textwidth]{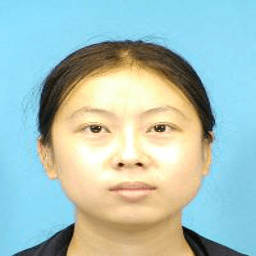}
		\includegraphics[width=0.145\textwidth]{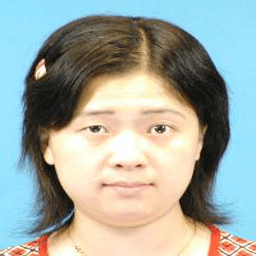}
		\includegraphics[width=0.145\textwidth]{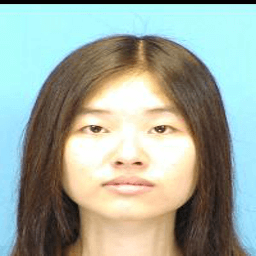}
		\includegraphics[width=0.145\textwidth]{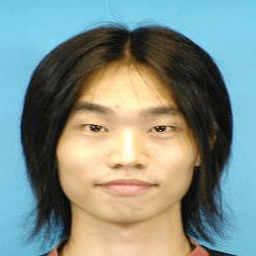}
		\includegraphics[width=0.145\textwidth]{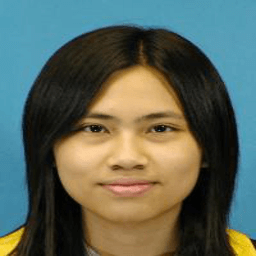}
	\end{subfigure}
	\begin{subfigure}[b]{0.45\textwidth}
		\centering
		\includegraphics[width=0.145\textwidth]{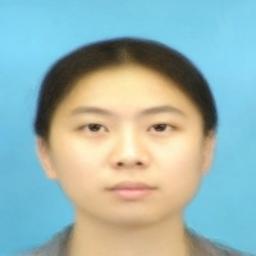}
		\includegraphics[width=0.145\textwidth]{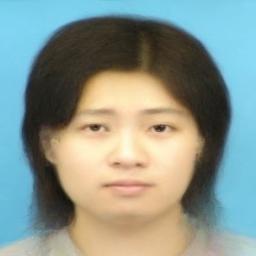}
		\includegraphics[width=0.145\textwidth]{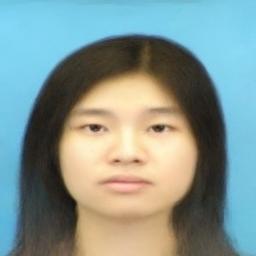}
		\includegraphics[width=0.145\textwidth]{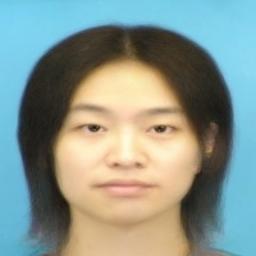}
		\includegraphics[width=0.145\textwidth]{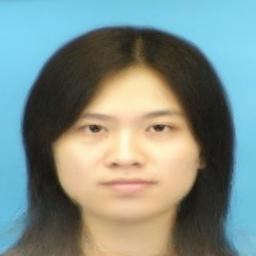}
	\end{subfigure}
	\begin{subfigure}[b]{0.45\textwidth}
		\centering
		\includegraphics[width=0.145\textwidth]{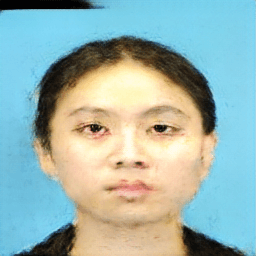}
		\includegraphics[width=0.145\textwidth]{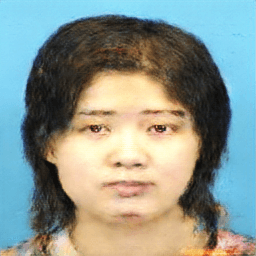}
		\includegraphics[width=0.145\textwidth]{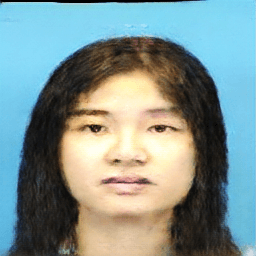}
		\includegraphics[width=0.145\textwidth]{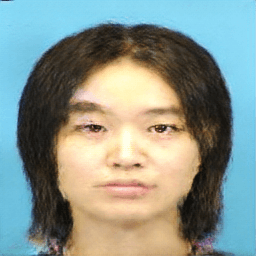}
		\includegraphics[width=0.145\textwidth]{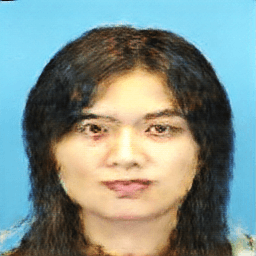}
	\end{subfigure}
	\begin{subfigure}[b]{0.45\textwidth}
		\centering
		\includegraphics[width=0.145\textwidth]{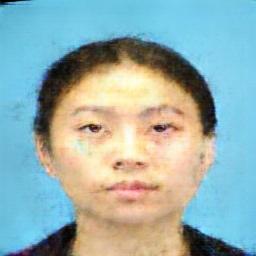}
		\includegraphics[width=0.145\textwidth]{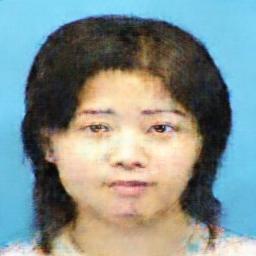}
		\includegraphics[width=0.145\textwidth]{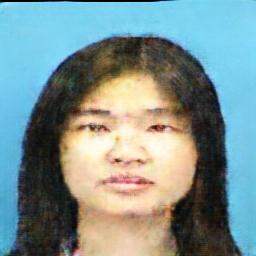}
		\includegraphics[width=0.145\textwidth]{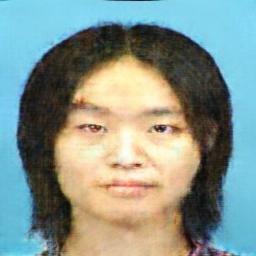}
		\includegraphics[width=0.145\textwidth]{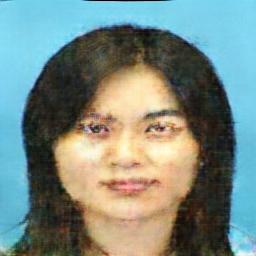}
	\end{subfigure}
	\begin{subfigure}[b]{0.45\textwidth}
		\centering
		\includegraphics[width=0.145\textwidth]{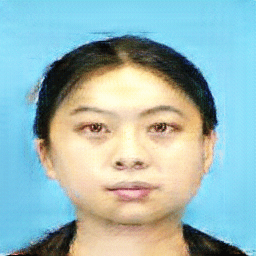}
		\includegraphics[width=0.145\textwidth]{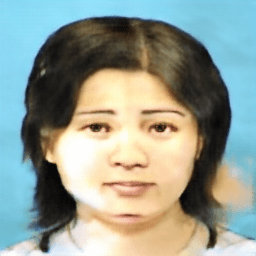}
		\includegraphics[width=0.145\textwidth]{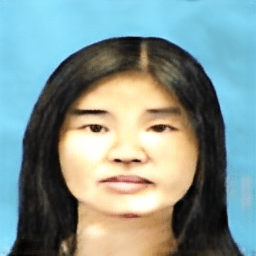}
		\includegraphics[width=0.145\textwidth]{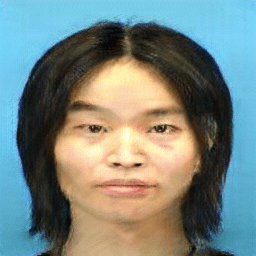}
		\includegraphics[width=0.145\textwidth]{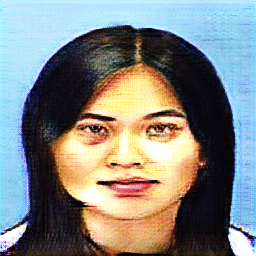}
	\end{subfigure}
	
	\begin{subfigure}[b]{0.45\textwidth}
		\centering
		\includegraphics[width=0.145\textwidth]{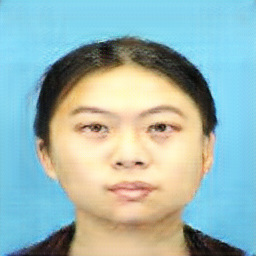}
		\includegraphics[width=0.145\textwidth]{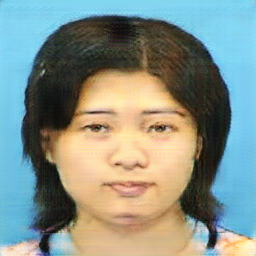}
		\includegraphics[width=0.145\textwidth]{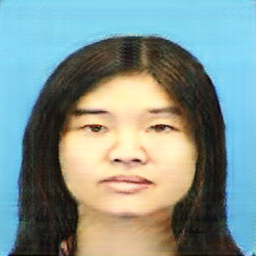}
		\includegraphics[width=0.145\textwidth]{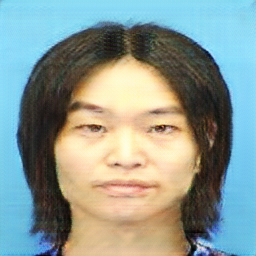}
		\includegraphics[width=0.145\textwidth]{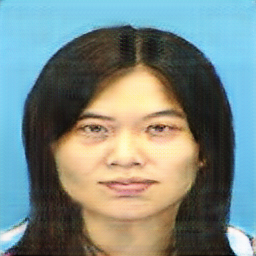}
	\end{subfigure}
	\caption{Comparison of Sketch to photo synthesis results on the CUHK dataset. From top to bottom: Input, Ground truth, MrFSPS, Pix2Pix, DualsAN, CycleGAN and PS\textsuperscript{2}-MAN. PS\textsuperscript{2}-MAN has minimal artifacts while generating realistic and sharper images. }\label{cuhkcompare2}
\end{figure}

\begin{figure}[t!]
	\centering
	\begin{subfigure}[b]{0.45\textwidth}
		\centering
		\includegraphics[width=0.145\textwidth]{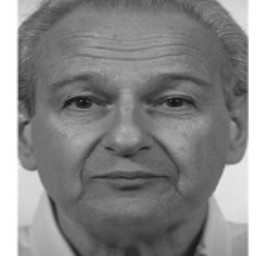}
		\includegraphics[width=0.145\textwidth]{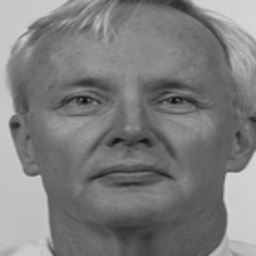}
		\includegraphics[width=0.145\textwidth]{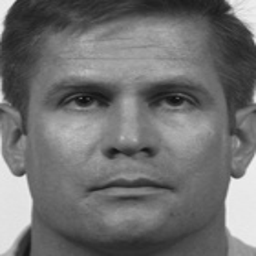}
		\includegraphics[width=0.145\textwidth]{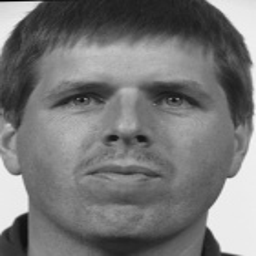}
		\includegraphics[width=0.145\textwidth]{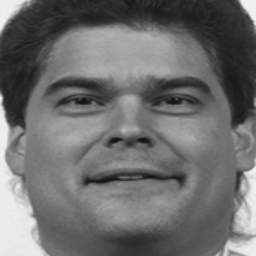}
	\end{subfigure}
	\begin{subfigure}[b]{0.45\textwidth}
		\centering
		\includegraphics[width=0.145\textwidth]{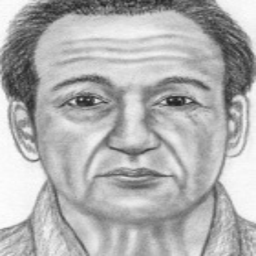}
		\includegraphics[width=0.145\textwidth]{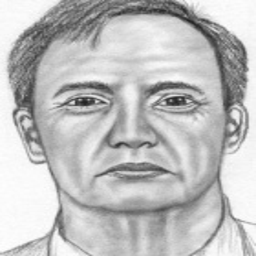}
		\includegraphics[width=0.145\textwidth]{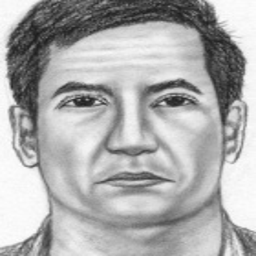}
		\includegraphics[width=0.145\textwidth]{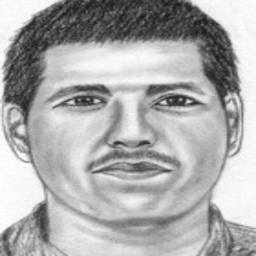}
		\includegraphics[width=0.145\textwidth]{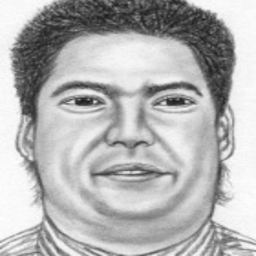}
	\end{subfigure}
	\begin{subfigure}[b]{0.45\textwidth}
		\centering
		\includegraphics[width=0.145\textwidth]{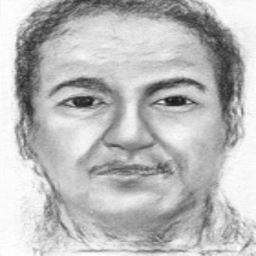}
		\includegraphics[width=0.145\textwidth]{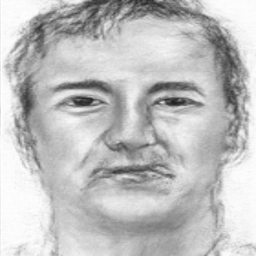}
		\includegraphics[width=0.145\textwidth]{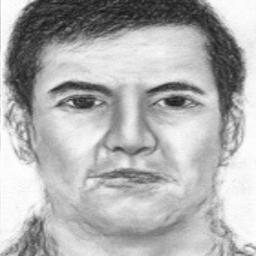}
		\includegraphics[width=0.145\textwidth]{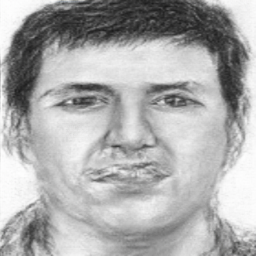}
		\includegraphics[width=0.145\textwidth]{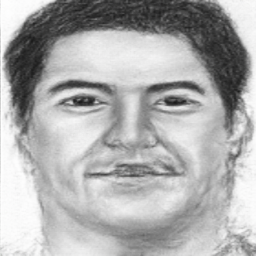}
	\end{subfigure}
	\begin{subfigure}[b]{0.45\textwidth}
		\centering
		\includegraphics[width=0.145\textwidth]{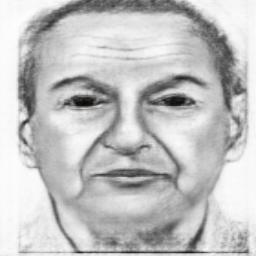}
		\includegraphics[width=0.145\textwidth]{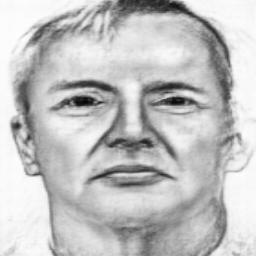}
		\includegraphics[width=0.145\textwidth]{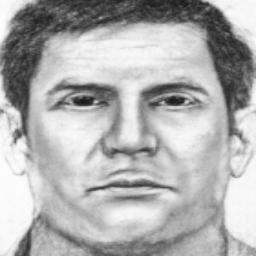}
		\includegraphics[width=0.145\textwidth]{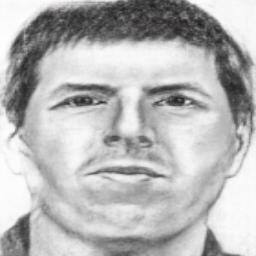}
		\includegraphics[width=0.145\textwidth]{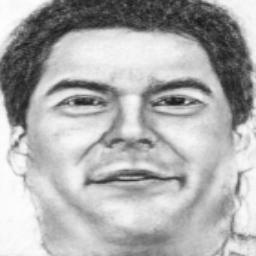}
	\end{subfigure}
	\begin{subfigure}[b]{0.45\textwidth}
		\centering
		\includegraphics[width=0.145\textwidth]{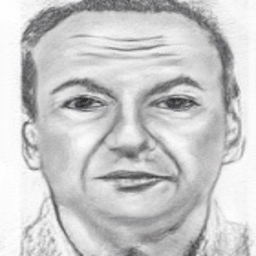}
		\includegraphics[width=0.145\textwidth]{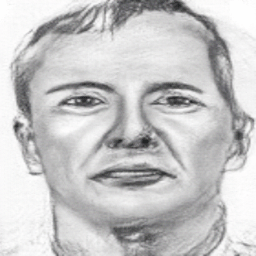}
		\includegraphics[width=0.145\textwidth]{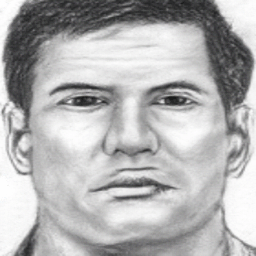}
		\includegraphics[width=0.145\textwidth]{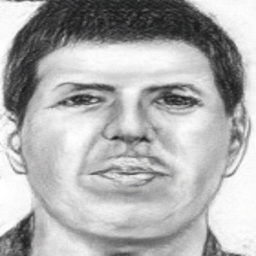}
		\includegraphics[width=0.145\textwidth]{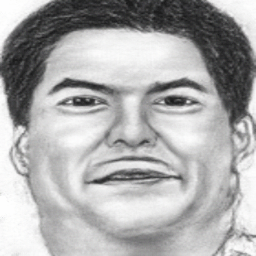}
	\end{subfigure}
	\begin{subfigure}[b]{0.45\textwidth}
		\centering
		\includegraphics[width=0.145\textwidth]{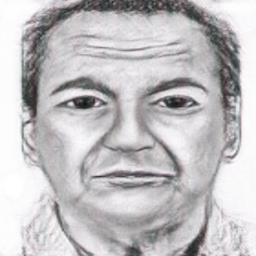}
		\includegraphics[width=0.145\textwidth]{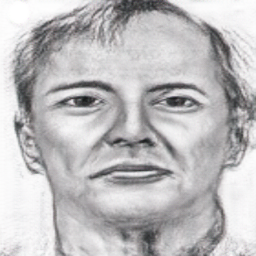}
		\includegraphics[width=0.145\textwidth]{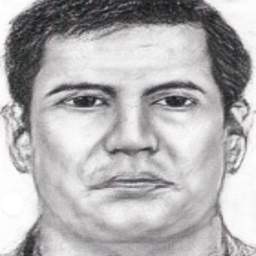}
		\includegraphics[width=0.145\textwidth]{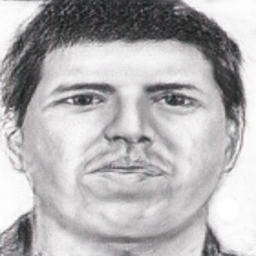}
		\includegraphics[width=0.145\textwidth]{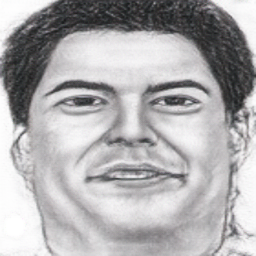}
	\end{subfigure}
	\caption{Comparison of photo to sketch synthesis results on the CUHK dataset. From top to bottom: Input, Ground truth, Pix2Pix, DualGAN, CycleGAN and PS\textsuperscript{2}-MAN. PS\textsuperscript{2}-MAN has minimal artifacts while generating realistic and sharper images.  }\label{cufsfcompare}
\end{figure}

\begin{figure}[t!]
	\centering
	\begin{subfigure}[b]{0.45\textwidth}
		\centering
		\includegraphics[width=0.145\textwidth]{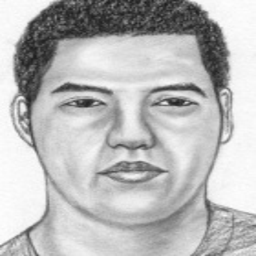}
		\includegraphics[width=0.145\textwidth]{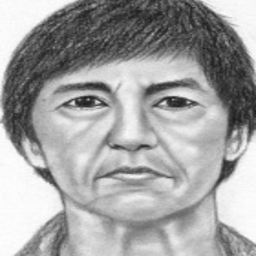}
		\includegraphics[width=0.145\textwidth]{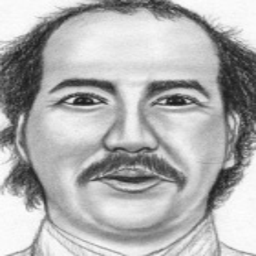}
		\includegraphics[width=0.145\textwidth]{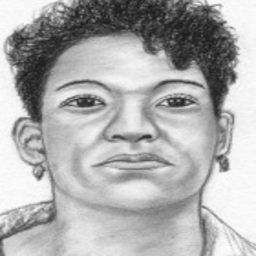}
		\includegraphics[width=0.145\textwidth]{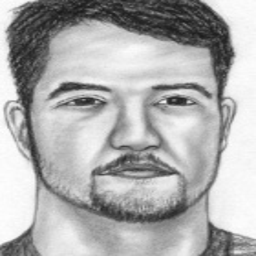}
	\end{subfigure}
	\begin{subfigure}[b]{0.45\textwidth}
		\centering
		\includegraphics[width=0.145\textwidth]{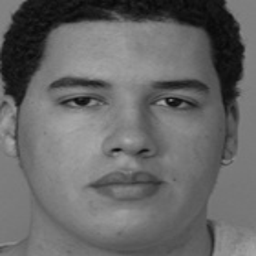}
		\includegraphics[width=0.145\textwidth]{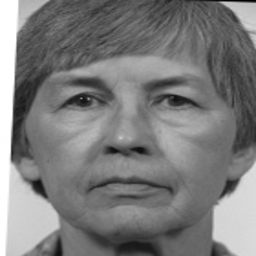}
		\includegraphics[width=0.145\textwidth]{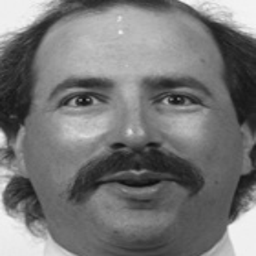}
		\includegraphics[width=0.145\textwidth]{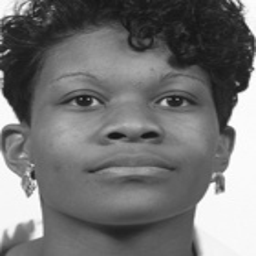}
		\includegraphics[width=0.145\textwidth]{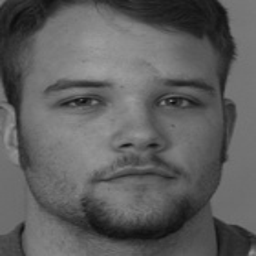}
	\end{subfigure}
	\begin{subfigure}[b]{0.45\textwidth}
		\centering
		\includegraphics[width=0.145\textwidth]{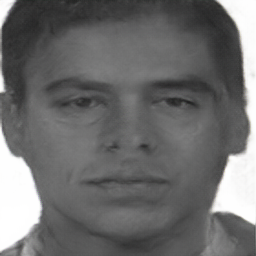}
		\includegraphics[width=0.145\textwidth]{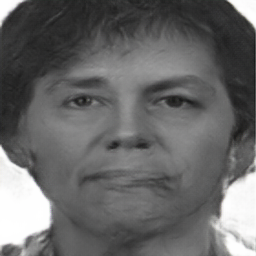}
		\includegraphics[width=0.145\textwidth]{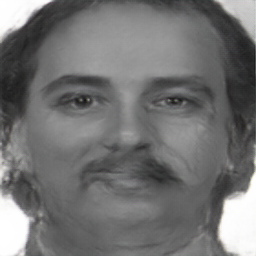}
		\includegraphics[width=0.145\textwidth]{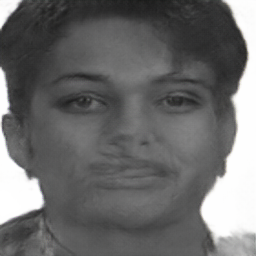}
		\includegraphics[width=0.145\textwidth]{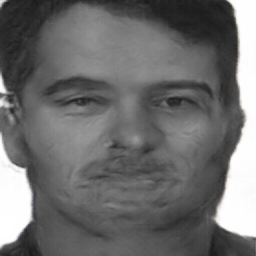}
	\end{subfigure}
	\begin{subfigure}[b]{0.45\textwidth}
		\centering
		\includegraphics[width=0.145\textwidth]{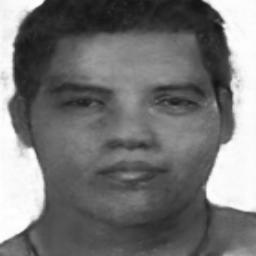}
		\includegraphics[width=0.145\textwidth]{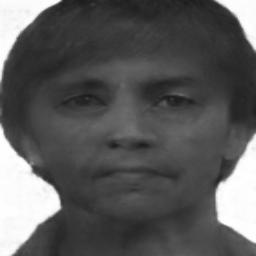}
		\includegraphics[width=0.145\textwidth]{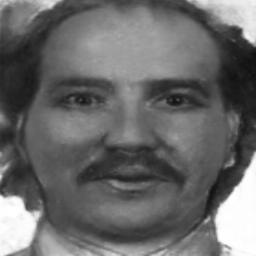}
		\includegraphics[width=0.145\textwidth]{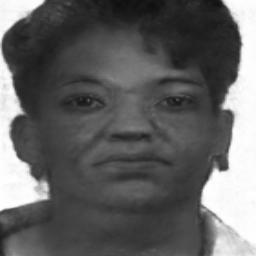}
		\includegraphics[width=0.145\textwidth]{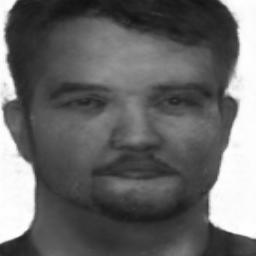}
	\end{subfigure}
	\begin{subfigure}[b]{0.45\textwidth}
		\centering
		\includegraphics[width=0.145\textwidth]{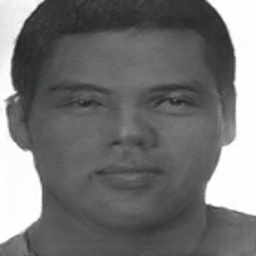}
		\includegraphics[width=0.145\textwidth]{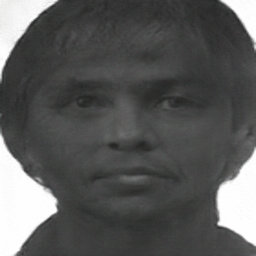}
		\includegraphics[width=0.145\textwidth]{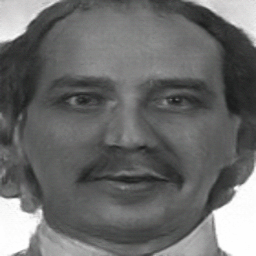}
		\includegraphics[width=0.145\textwidth]{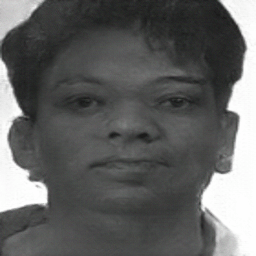}
		\includegraphics[width=0.145\textwidth]{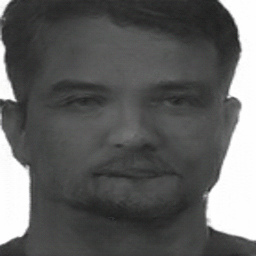}
	\end{subfigure}
	\begin{subfigure}[b]{0.45\textwidth}
		\centering
		\includegraphics[width=0.145\textwidth]{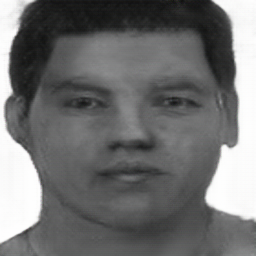}
		\includegraphics[width=0.145\textwidth]{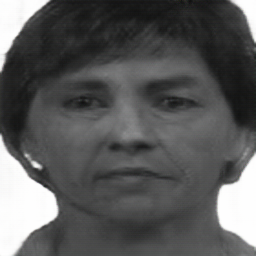}
		\includegraphics[width=0.145\textwidth]{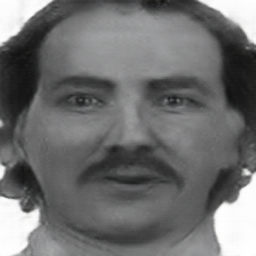}
		\includegraphics[width=0.145\textwidth]{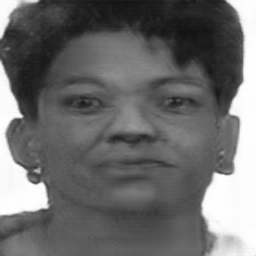}
		\includegraphics[width=0.145\textwidth]{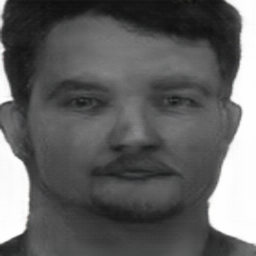}
	\end{subfigure}
	\caption{Comparison of sketch to photo synthesis results on the CUHK dataset. From top to bottom: Input, Ground truth, Pix2Pix, DualGAN, CycleGAN and PS\textsuperscript{2}-MAN. PS\textsuperscript{2}-MAN has minimal artifacts while generating realistic and sharper images. }\label{cufsfcompare2}
\end{figure}

\subsection{Comparison with state-of-the-art methods}

In addition to ablation studies, the proposed method is compared with recent state-of-the-art photo-sketch synthesis methods such as MWF \cite{zhou2012markov}, MrFSS \cite{peng2016multiple}, Pix2Pix \cite{isola2016image}, CycleGAN \cite{zhu2017unpaired} and DualGAN \cite{yi2017dualgan}. Sample sketch and photo synthesis results on the CUHK dataset are shown in Fig. \ref{cuhkcompare} and Fig. \ref{cuhkcompare2}, respectively. It can be observed that MrFSS synthesis results in blurred outputs. The generative models (Pix2Pix, CycleGAN and DualGAN) overcome the blurred effect by using  adversarial loss in addition to L1 loss. However, they tend to have undesirable artifacts due to instabilities in training while generating high-resolution images. In contrast, the proposed method (PS\textsuperscript{2}-MAN) is able to preserve high-frequency details and minimize the artifacts simultaneously. Also, photo synthesis using CycleGAN results in color distortion. A potential reason is the lack of L1 loss while training the network. Hence, in our case, we use L1 reconstruction error between target and synthesized image to train the network, thus providing the network with further regularization.

\begin{table}[t!]
	\centering
	\caption{\scriptsize{PERFORMANCE COMPARISON: QUANTITATIVE RESULTS FOR PHOTO AND SKETCH SYNTHESIS ON CUHK DATASET}}
	\label{compare}
	\vskip-5pt	\resizebox{0.48\textwidth}{!}{%
		\begin{tabular}{|l|c|c|c|c|c|c|}
			\hline
			& MWF    & MrFSPS & pix2pix & CycleGAN & DualGAN & Ours   \\ \hline
			SSIM (Photo Synthesis) & 0.6057 & 0.6326 & 0.6606  & 0.7626   & 0.7908  & 0.7915 \\ \hline
			SSIM (Sketch Synthesis) & 0.4996 & 0.5130 & 0.4669  & 0.5991   & 0.6003  & 0.6156 \\ \hline
			FSIM (Photo Synthesis) & 0.7996 & 0.8031 & 0.6997  & 0.7826   & 0.7939  & 0.8062 \\ \hline
			FSIM (Sketch Synthesis) & 0.7121 & 0.7339 & 0.6174  & 0.7271   & 0.7312  & 0.7361 \\ \hline
		\end{tabular}
	}
\end{table}  

Sample sketch and photo synthesis results on the CUFSF dataset for the generative techniques are shown in Fig. \ref{cufsfcompare} and Fig. \ref{cufsfcompare2}, respectively. The CUFSF dataset is particularly challenging since the sketches have over-exaggerated features as compared to the ones present in the real photos.  It can be observed that in case of both sketch and photo synthesis that the generative methods (Pix2Pix, CycleGAN and DualGAN) introduce undesirable artifacts especially at facial features resulting.
In contrast, the proposed method is able to minimize the artifacts while generating realistic images as compared to the other methods. 

\begin{figure}[ht!]
	\centering
	\includegraphics[width=0.23\textwidth]{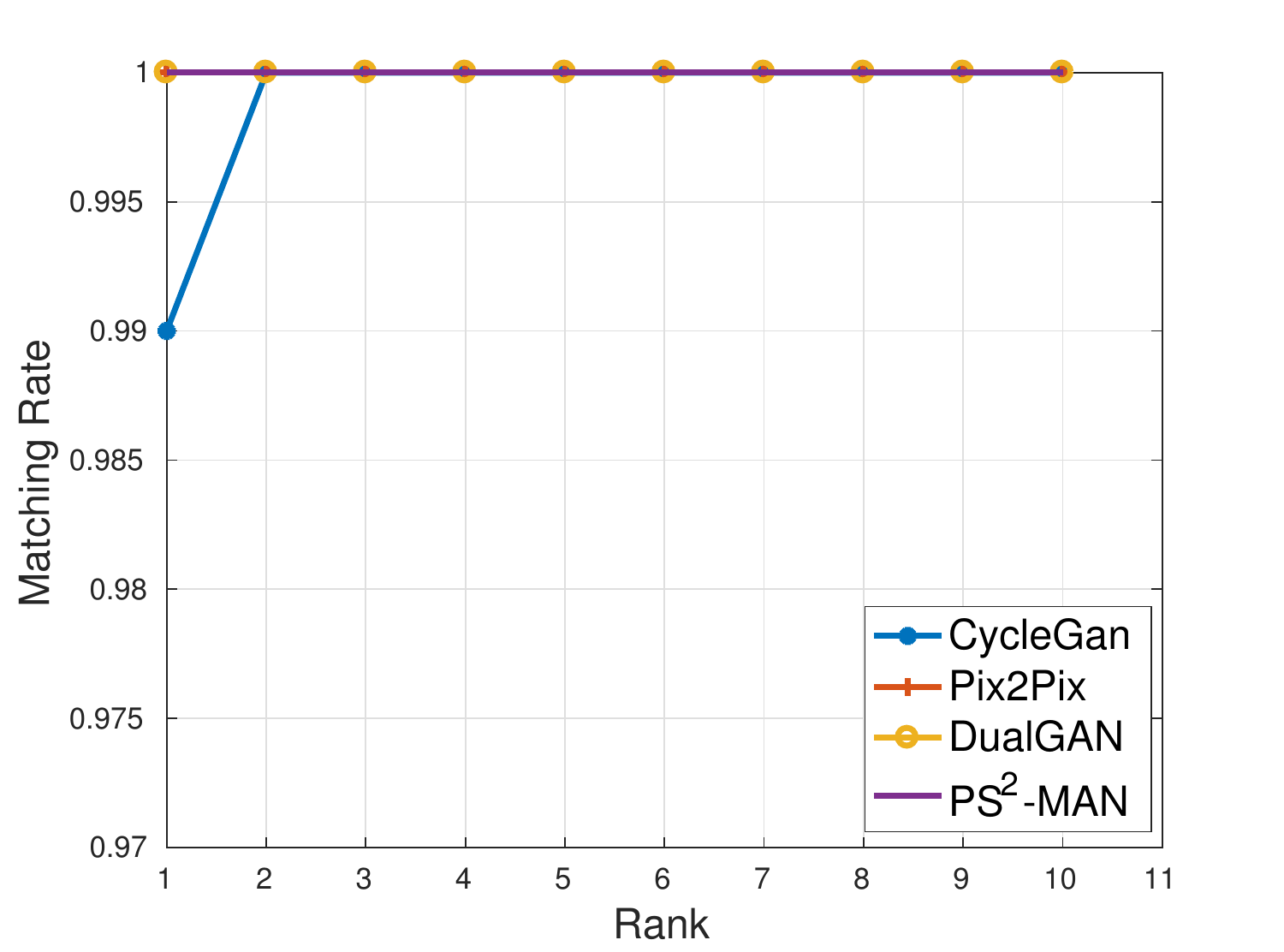}
	\includegraphics[width=0.23\textwidth]{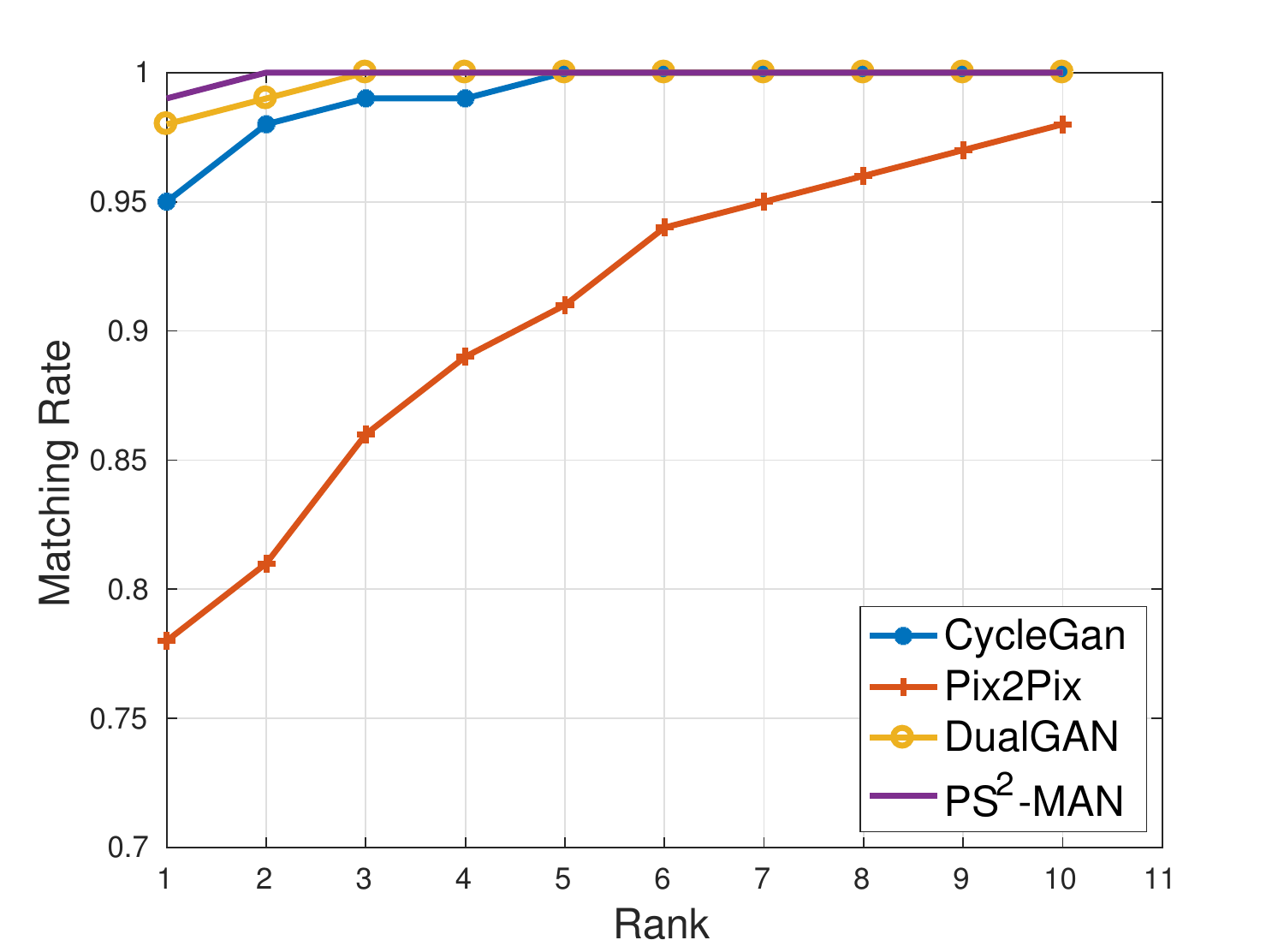} \\
	(a)\hskip110pt(b)
	\vskip-5pt
	\caption{Matching rates using generative techniques on CUHK dataset for different ranks (a) Photo matching rates (b) Sketch matching rates.}\label{fig:matching_CUFS}
\end{figure}

\begin{figure}[ht!]
	\centering
	\includegraphics[width=0.23\textwidth]{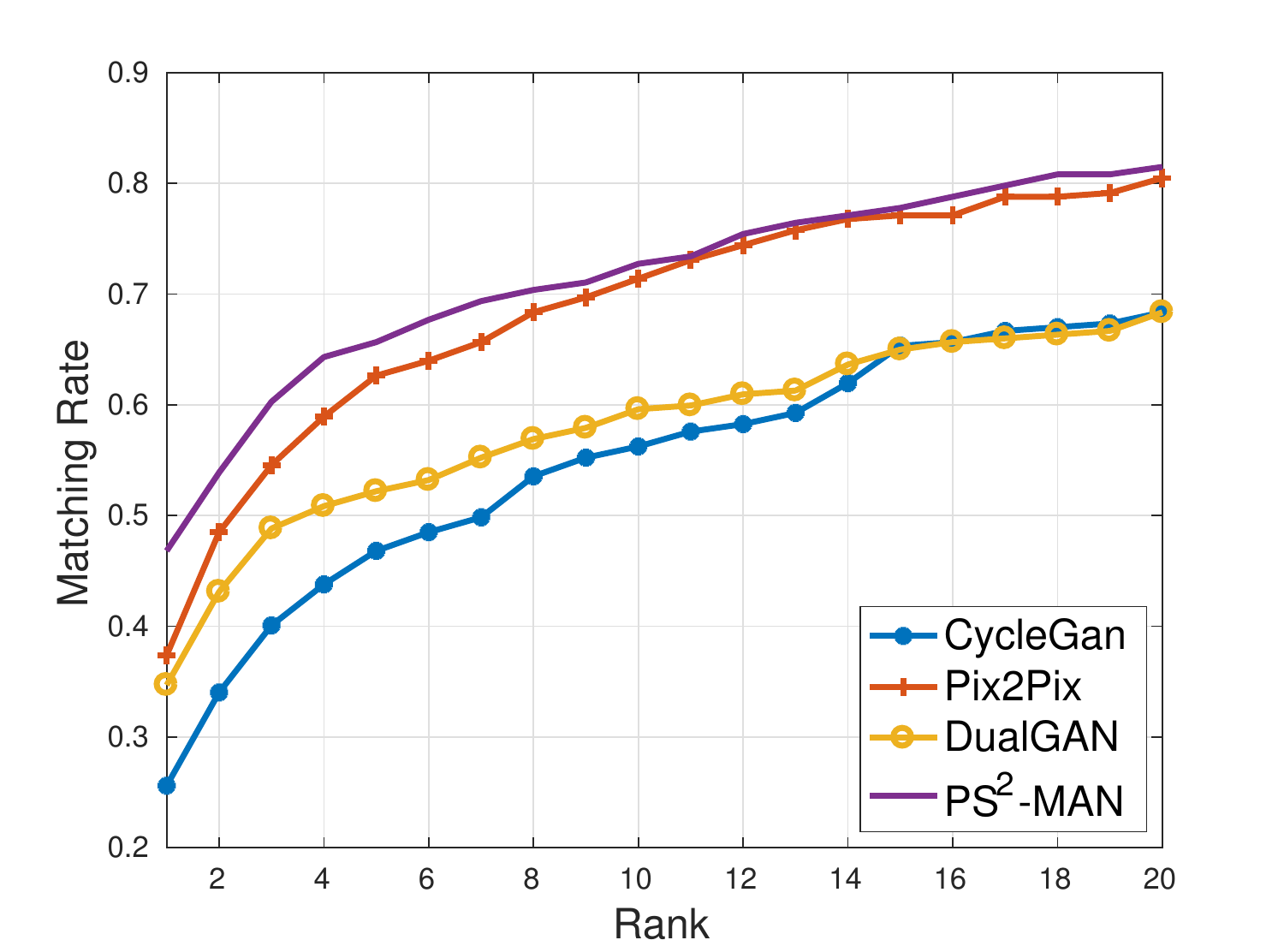}
	\includegraphics[width=0.23\textwidth]{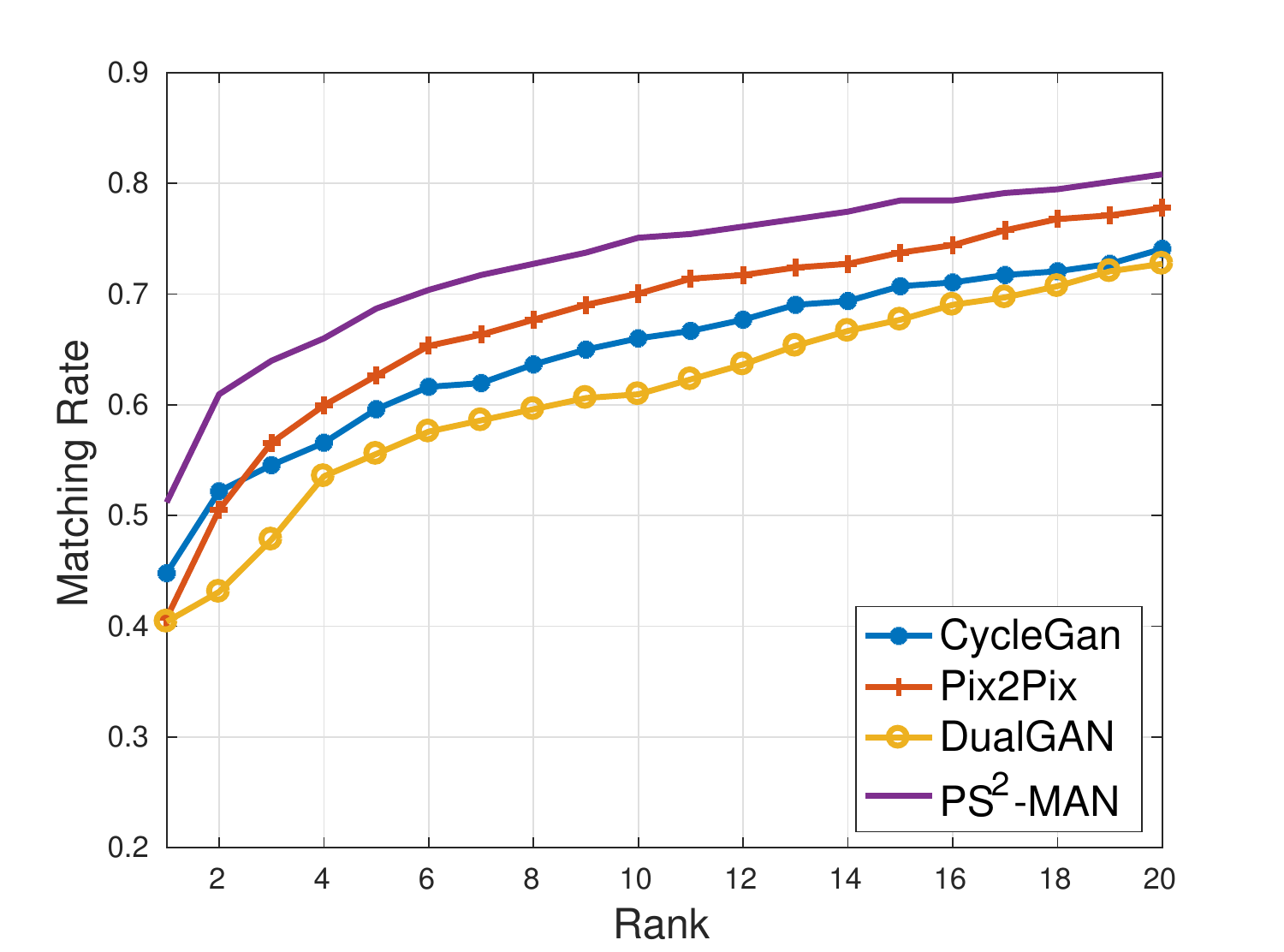}
	(a)\hskip110pt(b)
	\vskip-5pt
	\caption{Matching rates using generative techniques on CUFSF dataset for different ranks (a) Photo matching rates (b) Sketch matching rates.}\label{fig:matching_cufsf}
\end{figure}

Similar to ablation studies, we also compare the results of all the above methods using quantitative measures (SSIM and FSIM) as shown in Table \ref{compare}.  The proposed method achieves the best results in terms of SSIM and FSIM as compared to the other methods. Additionally, the methods are also compared using photo-sketch face  matching rates using two approaches: (1) Synthesize sketches from photos and used these synthesized sketches to match with real sketch gallery. (2) Synthesize photos from sketches and use these synthesized photos to match with real photos gallery. The matching rates were calculated by computing the LBP features and cosine distance.  The matching rates using generative techniques on the CUHK and CUFSF datasets for various are illustrated in Fig. \ref{fig:matching_CUFS} and \ref{fig:matching_cufsf} and respectively in terms of the Cumulative Matching Characteristic (CMC) curves.  Table \ref{matching}  summarize the rank-1 matching rates. It can be observed from Fig. \ref{fig:matching_CUFS} and \ref{fig:matching_cufsf} that the proposed method achieves best matching rates at all ranks. 

To summarize, through various experiments it is demonstrated that the proposed method PS\textsuperscript{2}-MAN is able to generate realistic results with minimal artifacts as compared to existing methods.  This is mainly due to the multi-adversarial network used in our approach.   Additionally, the proposed method achieves significant improvements over the other techniques in terms of various quality measures (such as SSIM and FSIM ) and matching rates while generating visually appealing outputs.

\begin{table}[h!]
	\centering
	\caption{\scriptsize{RANK-1 MATCHING RATES FOR GENERATIVE METHODS ON CUHK AND CUFSF DATASETS}}
	\label{matching}
	\resizebox{0.48\textwidth}{!}{%
		\begin{tabular}{|l|c|c|c|c|c|}
			\hline
			Dataset                & Photo/Sketch    & Pix2Pix & CycleGAN & DualGAN & PS\textsuperscript{2}-MAN \\ \hline
			\multirow{2}{*}{CUHK}  & Photo Matching  & 100     & 99       & 100     & 100     \\ \cline{2-6} 
			& Sketch Matching & 78      & 95       & 98      & 99      \\ \hline
			\multirow{2}{*}{CUFSF} & Photo Matching  & 37      & 25       & 35      & 47      \\ \cline{2-6} 
			& Sketch Matching & 40      & 44       & 40      & 51      \\ \hline
		\end{tabular}
	}
\end{table}

\section{CONCLUSION}

We explored the problem of photo-sketch synthesis using the recently introduced generative models. A novel synthesis method using multi-adversarial networks is presented. The proposed method is developed specifically to enable GANs to generate high resolution images. This is achieved by providing adversarial supervision to hidden layers of the generator sub-network. Additionally, the forward and backward synthesis are trained iteratively in the CycleGAN framework, i.e., in addition to minimizing L1 reconstruction error, cycle-consistency loss is also used in the objective function. These additional loss functions provide appropriate regularization thereby generating high-quality and high resolution photo-sketch synthesis. Evaluations are performed on two popular datasets and the results are compared with recent state-of-the-art generative methods. It is clearly demonstrated that the proposed method achieves significant improvements in terms of visual quality and photo/sketch matching rates.

\section*{Acknowledgment}
This research is based upon work supported by the Office of the Director
of  National  Intelligence  (ODNI),  Intelligence  Advanced  Research  Projects
Activity  (IARPA),  via  IARPA  R\&D  Contract  No.  2014-14071600012.  The
views and conclusions contained herein are those of the authors and should not
be interpreted as necessarily representing the official policies or endorsements,
either  expressed  or  implied,  of  the  ODNI,  IARPA,  or  the  U.S.  Government.
The  U.S.  Government  is  authorized  to  reproduce  and  distribute  reprints  for
Governmental purposes notwithstanding any copyright annotation thereon.   
%


{\small\bibliographystyle{ieee}
	\bibliography{sketch}

\begin{thebibliography}{10}\itemsep=-1pt

\bibitem{chang2010face}
L.~Chang, M.~Zhou, Y.~Han, and X.~Deng.
\newblock Face sketch synthesis via sparse representation.
\newblock In {\em 20th ICPR}, pages 2146--2149. IEEE, 2010.

\bibitem{chen2018face}
C.~Chen, X.~Tax, and K.~Wong.
\newblock Face sketch synthesis with style transfer using pyramid column
  feature.
\newblock In {\em IEEE WACV}, 2018.

\bibitem{di2017face}
X.~Di and V.~M. Patel.
\newblock Face synthesis from visual attributes via sketch using conditional
  vaes and gans.
\newblock {\em arXiv:1801.00077}, 2017.

\bibitem{di2017gp}
X.~Di, V.~A. Sindagi, and V.~M. Patel.
\newblock Gp-gan: Gender preserving gan for synthesizing faces from landmarks.
\newblock {\em arXiv:1710.00962}, 2017.

\bibitem{gao2017composition}
F.~Gao, S.~Shi, J.~Yu, and Q.~Huang.
\newblock Composition-aided sketch-realistic portrait generation.
\newblock {\em arXiv:1712.00899}, 2017.

\bibitem{gao2012face}
X.~Gao, N.~Wang, D.~Tao, and X.~Li.
\newblock Face sketch--photo synthesis and retrieval using sparse
  representation.
\newblock {\em IEEE CSVT}, 22(8):1213--1226, 2012.

\bibitem{gao2008face}
X.~Gao, J.~Zhong, J.~Li, and C.~Tian.
\newblock Face sketch synthesis algorithm based on e-hmm and selective
  ensemble.
\newblock {\em IEEE CSVT}, 18(4):487--496, 2008.

\bibitem{goodfellow2014generative}
I.~Goodfellow, J.~Pouget-Abadie, M.~Mirza, B.~Xu, D.~Warde-Farley, S.~Ozair,
  A.~Courville, and Y.~Bengio.
\newblock Generative adversarial nets.
\newblock In {\em Advances in NIPS}, pages 2672--2680, 2014.

\bibitem{han2013matching}
H.~Han, B.~F. Klare, K.~Bonnen, and A.~K. Jain.
\newblock Matching composite sketches to face photos: A component-based
  approach.
\newblock {\em IEEE IFS}, 8(1):191--204, 2013.

\bibitem{he2016deep}
K.~He, X.~Zhang, S.~Ren, and J.~Sun.
\newblock Deep residual learning for image recognition.
\newblock In {\em IEEE CVPR}, pages 770--778, 2016.

\bibitem{isola2016image}
P.~Isola, J.-Y. Zhu, T.~Zhou, and A.~A. Efros.
\newblock Image-to-image translation with conditional adversarial networks.
\newblock {\em arXiv:1611.07004}, 2016.

\bibitem{kingma2014adam}
D.~Kingma and J.~Ba.
\newblock Adam: A method for stochastic optimization.
\newblock {\em arXiv:1412.6980}, 2014.

\bibitem{kingma2013auto}
D.~P. Kingma and M.~Welling.
\newblock Auto-encoding variational bayes.
\newblock {\em arXiv:1312.6114}, 2013.

\bibitem{klare2011matching}
B.~Klare, Z.~Li, and A.~K. Jain.
\newblock Matching forensic sketches to mug shot photos.
\newblock {\em IEEE PAMI}, 33(3):639--646, 2011.

\bibitem{klare2013heterogeneous}
B.~F. Klare and A.~K. Jain.
\newblock Heterogeneous face recognition using kernel prototype similarities.
\newblock {\em IEEE PAMI}, 35(6):1410--1422, 2013.

\bibitem{liu2005nonlinear}
Q.~Liu, X.~Tang, H.~Jin, H.~Lu, and S.~Ma.
\newblock A nonlinear approach for face sketch synthesis and recognition.
\newblock In {\em CVPR 2005}, volume~1, pages 1005--1010. IEEE, 2005.

\bibitem{martinez1998ar}
A.~Martinez and R.~Benavente.
\newblock The ar face database, cvc.
\newblock 1998.

\bibitem{messer1999xm2vtsdb}
K.~Messer, J.~Matas, J.~Kittler, J.~Luettin, and G.~Maitre.
\newblock Xm2vtsdb: The extended m2vts database.
\newblock In {\em Second international conference on audio and video-based
  biometric person authentication}, volume 964, pages 965--966, 1999.

\bibitem{ouyang2016forgetmenot}
S.~Ouyang, T.~M. Hospedales, Y.-Z. Song, and X.~Li.
\newblock Forgetmenot: memory-aware forensic facial sketch matching.
\newblock In {\em IEEE CVPR}, pages 5571--5579, 2016.

\bibitem{peng2017superpixel}
C.~Peng, X.~Gao, N.~Wang, and J.~Li.
\newblock Superpixel-based face sketch--photo synthesis.
\newblock {\em IEEE CSVT}, 27(2):288--299, 2017.

\bibitem{peng2016multiple}
C.~Peng, X.~Gao, N.~Wang, D.~Tao, X.~Li, and J.~Li.
\newblock Multiple representations-based face sketch--photo synthesis.
\newblock {\em IEEE NNLS}, 27(11):2201--2215, 2016.

\bibitem{peng2016face}
C.~Peng, N.~Wang, X.~Gao, and J.~Li.
\newblock Face recognition from multiple stylistic sketches: Scenarios,
  datasets, and evaluation.
\newblock In {\em ECCV}, 2016.

\bibitem{peng2015piefa}
X.~Peng, S.~Zhang, Y.~Yang, and D.~N. Metaxas.
\newblock Piefa: Personalized incremental and ensemble face alignment.
\newblock In {\em IEEE ICCV}, pages 3880--3888, 2015.

\bibitem{perera2017in2i}
P.~Perera, M.~Abavisani, and V.~M. Patel.
\newblock In2i: Unsupervised multi-image-to-image translation using generative
  adversarial networks.
\newblock {\em arXiv:1711.09334}, 2017.

\bibitem{phillips1998feret}
P.~J. Phillips, H.~Wechsler, J.~Huang, and P.~J. Rauss.
\newblock The feret database and evaluation procedure for face-recognition
  algorithms.
\newblock {\em Image and vision computing}, 16(5):295--306, 1998.

\bibitem{rezende2014stochastic}
D.~J. Rezende, S.~Mohamed, and D.~Wierstra.
\newblock Stochastic backpropagation and approximate inference in deep
  generative models.
\newblock {\em arXiv:1401.4082}, 2014.

\bibitem{sangkloy2016scribbler}
P.~Sangkloy, J.~Lu, C.~Fang, F.~Yu, and J.~Hays.
\newblock Scribbler: Controlling deep image synthesis with sketch and color.
\newblock {\em arXiv:1612.00835}, 2016.

\bibitem{sindagi2017generating}
V.~A. Sindagi and V.~M. Patel.
\newblock Generating high-quality crowd density maps using contextual pyramid
  cnns.
\newblock In {\em IEEE ICCV}, 2017.

\bibitem{tang2002face}
X.~Tang and X.~Wang.
\newblock Face photo recognition using sketch.
\newblock In {\em ICIP}, volume~1, pages I--I. IEEE, 2002.

\bibitem{tang2004face}
X.~Tang and X.~Wang.
\newblock Face sketch recognition.
\newblock {\em CSVT}, 14(1):50--57, 2004.

\bibitem{wang2017bayesian}
N.~Wang, X.~Gao, L.~Sun, and J.~Li.
\newblock Bayesian face sketch synthesis.
\newblock {\em IEEE TIP}, 26(3):1264--1274, 2017.

\bibitem{wang2013heterogeneous}
N.~Wang, J.~Li, D.~Tao, X.~Li, and X.~Gao.
\newblock Heterogeneous image transformation.
\newblock {\em Pattern Recognition Letters}, 34(1):77--84, 2013.

\bibitem{wang2013transductive}
N.~Wang, D.~Tao, X.~Gao, X.~Li, and J.~Li.
\newblock Transductive face sketch-photo synthesis.
\newblock {\em IEEE NNLS}, 24(9):1364--1376, 2013.

\bibitem{wang2014comprehensive}
N.~Wang, D.~Tao, X.~Gao, X.~Li, and J.~Li.
\newblock A comprehensive survey to face hallucination.
\newblock {\em IJCV}, 106(1):9--30, 2014.

\bibitem{wang2009face}
X.~Wang and X.~Tang.
\newblock Face photo-sketch synthesis and recognition.
\newblock {\em TPAMI}, 31(11):1955--1967, 2009.

\bibitem{wang2004image}
Z.~Wang, A.~C. Bovik, H.~R. Sheikh, and E.~P. Simoncelli.
\newblock Image quality assessment: from error visibility to structural
  similarity.
\newblock {\em IEEE TIP}, 13(4):600--612, 2004.

\bibitem{wolf2017unsupervised}
L.~Wolf, Y.~Taigman, and A.~Polyak.
\newblock Unsupervised creation of parameterized avatars.
\newblock {\em arXiv:1704.05693}, 2017.

\bibitem{xiao2009new}
B.~Xiao, X.~Gao, D.~Tao, and X.~Li.
\newblock A new approach for face recognition by sketches in photos.
\newblock {\em Signal Processing}, 89(8):1576--1588, 2009.

\bibitem{yi2017dualgan}
Z.~Yi, H.~Zhang, P.~T. Gong, et~al.
\newblock Dualgan: Unsupervised dual learning for image-to-image translation.
\newblock {\em IEEE ICCV}, 2017.

\bibitem{zhang2017image}
H.~Zhang, V.~Sindagi, and V.~M. Patel.
\newblock Image de-raining using a conditional generative adversarial network.
\newblock {\em arXiv:1701.05957}, 2017.

\bibitem{zhang2017joint}
H.~Zhang, V.~Sindagi, and V.~M. Patel.
\newblock Joint transmission map estimation and dehazing using deep networks.
\newblock {\em arXiv:1708.00581}, 2017.

\bibitem{zhang2016stackgan}
H.~Zhang, T.~Xu, H.~Li, S.~Zhang, X.~Huang, X.~Wang, and D.~Metaxas.
\newblock Stackgan: Text to photo-realistic image synthesis with stacked
  generative adversarial networks.
\newblock In {\em IEEE ICCV}, 2017.

\bibitem{zhang2015end}
L.~Zhang, L.~Lin, X.~Wu, S.~Ding, and L.~Zhang.
\newblock End-to-end photo-sketch generation via fully convolutional
  representation learning.
\newblock In {\em ACM ICMR}, pages 627--634. ACM, 2015.

\bibitem{zhang2011fsim}
L.~Zhang, L.~Zhang, X.~Mou, and D.~Zhang.
\newblock Fsim: A feature similarity index for image quality assessment.
\newblock {\em IEEE TIP}, 20(8):2378--2386, 2011.

\bibitem{zhang2016robust}
S.~Zhang, X.~Gao, N.~Wang, and J.~Li.
\newblock Robust face sketch style synthesis.
\newblock {\em IEEE TIP}, 25(1):220--232, 2016.

\bibitem{zhang2011coupled}
W.~Zhang, X.~Wang, and X.~Tang.
\newblock Coupled information-theoretic encoding for face photo-sketch
  recognition.
\newblock In {\em 2011 IEEE CVPR}, pages 513--520. IEEE, 2011.

\bibitem{zheng2017photo}
Z.~Zheng, H.~Zheng, Z.~Yu, Z.~Gu, and B.~Zheng.
\newblock Photo-to-caricature translation on faces in the wild.
\newblock {\em arXiv:1711.10735}, 2017.

\bibitem{zhou2012markov}
H.~Zhou, Z.~Kuang, and K.-Y.~K. Wong.
\newblock Markov weight fields for face sketch synthesis.
\newblock In {\em CVPR}, pages 1091--1097. IEEE, 2012.

\bibitem{zhu2017unpaired}
J.-Y. Zhu, T.~Park, P.~Isola, and A.~A. Efros.
\newblock Unpaired image-to-image translation using cycle-consistent
  adversarial networks.
\newblock {\em IEEE ICCV}, 2017.

\end{thebibliography}
}

\end{document}